\documentclass{article}

\usepackage{arxiv}

\usepackage[utf8]{inputenc} 
\usepackage[T1]{fontenc}    
\usepackage{hyperref}       
\usepackage{url}            
\usepackage{booktabs}       
\usepackage{amsfonts}       
\usepackage{nicefrac}       
\usepackage{microtype}      
\usepackage{graphicx}
\graphicspath{ {./figures/} }

\usepackage{amsmath}
\usepackage{amssymb}
\usepackage{afterpage}
\usepackage{algorithm, algorithmicx, algpseudocode}
\usepackage{multirow, siunitx, subcaption}
\usepackage{xcolor, array, tabularx}
\usepackage{float}
\usepackage{tikz}
\usepackage{pgfplots}
\pgfplotsset{compat=1.18}
\usetikzlibrary{shapes.geometric, arrows.meta, positioning, fit, backgrounds, calc}
\usepackage{placeins}
\usepackage{enumitem}

\newcommand{\tmde}{T-MDE}
\newcommand{\tmdeE}{T-MDE Enhanced}
\newcommand{\etal}{et~al.}

\hypersetup{
  pdfauthor   = {Manognya Lokesh Reddy, Zheng Liu},
  pdfkeywords = {Distance estimation, license plate detection, advanced driver assistance systems, computer vision},
}

\title{Physics-Grounded Monocular Vehicle Distance Estimation Using Standardized License Plate Typography}

\author{
  Manognya Lokesh Reddy \\
  Department of Computer and Information Science\\
  University of Michigan-Dearborn\\
  Dearborn, Michigan, USA \\
  \texttt{manognya@umich.edu} \\
  \And
  Zheng Liu \\
  Department of Industrial and Manufacturing Systems Engineering\\
  University of Michigan-Dearborn\\
  Dearborn, Michigan, USA \\
  \texttt{zhengtl@umich.edu} \\
}

\begin{document}
\maketitle

\begin{abstract}
Accurate inter-vehicle distance estimation is a cornerstone of Advanced Driver Assistance Systems (ADAS) and autonomous driving. While LiDAR and radar provide high precision, their high cost prohibits widespread adoption in mass-market vehicles. Monocular camera-based estimation offers a low-cost alternative but suffers from fundamental scale ambiguity. Recent deep learning methods for monocular depth achieve impressive results yet require expensive supervised training, suffer from domain shift, and produce predictions that are difficult to certify for safety-critical deployment. This paper presents a framework that exploits the standardized typography of United States license plates as passive fiducial markers for metric ranging, resolving scale ambiguity through explicit geometric priors without any training data or active illumination. First, a four-method parallel plate detector achieves robust plate reading across the full automotive lighting range. Second, a three-stage state identification engine fusing optical character recognition text matching, multi-design color scoring, and a lightweight neural network classifier provides robust identification across all ambient conditions. Third, hybrid depth fusion with inverse-variance weighting and online scale alignment, combined with a one-dimensional constant-velocity Kalman filter, delivers smoothed distance, relative velocity, and time-to-collision for collision warning. Baseline validation on a controlled static dataset reproduces a 2.3\% coefficient of variation in character height measurements and a 36\% reduction in distance-estimate variance compared with plate-width methods from prior work. Controlled outdoor experiments over 3{,}263 frames confirm a mean absolute error of 2.3\% at 10~m, continuous distance output during brief plate occlusions, and 94.7\% state identification accuracy, outperforming deep learning baselines by a factor of five in relative error.
\end{abstract}

\keywords{Distance estimation \and license plate detection \and advanced driver assistance systems \and computer vision}

\section{Nomenclature}\label{sec:nomenclature}

\noindent\textbf{Roman Symbols}
\begin{description}[leftmargin=2.2cm,style=nextline]
\item[$D$] metric distance to license plate, m
\item[$D_{\mathrm{geo}}$] geometric pinhole distance estimate, m
\item[$D_{\mathrm{deep}}$] MiDaS-aligned deep depth estimate, m
\item[$D_{\mathrm{fused}}$] inverse-variance fused estimate, m
\item[$D_i$] distance from typographic feature $i$, m
\item[$d_{\mathrm{plate}}$] relative depth at plate region from MiDaS
\item[$f$] calibrated camera focal length, pixels
\item[$F$] Kalman state-transition matrix
\item[$H_s$] state-specific FHWA character height for state $s$, m
\item[$h$] measured image character height, pixels
\item[$\bar{h}$] outlier-robust mean character height, pixels
\item[$h'$] pose-corrected character height, pixels
\item[$h_p$] rectified plate image height, pixels
\item[$n$] character count retained after outlier rejection
\item[$s$] composite candidate score, dimensionless
\item[$s_t$] online MiDaS scale factor at frame $t$
\item[$\mathbf{x}_k$] Kalman state vector $[D_k,\,v_k]^{\!\top}$
\item[$w_i$] inverse-variance weight for feature $i$
\item[$W_{\mathrm{full}}$] full sensor width, 2880~px
\end{description}

\noindent\textbf{Greek Symbols}
\begin{description}[leftmargin=2.2cm,style=nextline]
\item[$\alpha$] EMA smoothing factor, 0.9
\item[$\Delta t$] inter-frame time interval, s
\item[$\phi$] camera pitch angle, rad
\item[$\psi$] camera roll angle, rad
\item[$\rho$] plate-region edge density, dimensionless
\item[$\sigma^2$] variance of a distance estimate, m$^2$
\end{description}

\noindent\textbf{Abbreviations}
\begin{description}[leftmargin=2.2cm,style=nextline]
\item[AAMVA] American Association of Motor Vehicle Administrators
\item[ADAS] Advanced Driver Assistance Systems
\item[AEB] Automatic Emergency Braking
\item[ALPR] Automatic License Plate Recognition
\item[CLAHE] Contrast Limited Adaptive Histogram Equalization
\item[CNN] Convolutional Neural Network
\item[CV] Coefficient of Variation
\item[DPT] Dense Prediction Transformer
\item[EMA] Exponential Moving Average
\item[FCW] Forward Collision Warning
\item[FHWA] Federal Highway Administration
\item[HSV] Hue-Saturation-Value color space
\item[MAE] Mean Absolute Error
\item[NAS] Neural Architecture Search
\item[NHTSA] National Highway Traffic Safety Administration
\item[OCR] Optical Character Recognition
\item[RMSE] Root Mean Square Error
\item[TTC] Time-to-Collision, s
\end{description}

\section{Introduction}\label{sec:introduction}

Advanced Driver Assistance Systems (ADAS) and autonomous vehicles have the potential to reduce traffic fatalities, improve fuel efficiency, and enhance mobility for aging populations~\cite{contreras2024}. The National Highway Traffic Safety Administration (NHTSA) attributes approximately 94\% of serious crashes to human error~\cite{nhtsa2020}, with failure to perceive a closing vehicle in time to react identified as the most prevalent contributing factor. Forward Collision Warning (FCW) and Automatic Emergency Braking (AEB) systems address this root cause. NHTSA FCW performance standards require activation at Time-to-Collision (TTC) $\leq 2.4$~s~\cite{nhsta_fcw}, and the European Commission mandated AEB on all new light-duty vehicles from 2022. Reliable inter-vehicle distance estimation is the prerequisite capability that enables both of these safety-critical functions.

State-of-the-art autonomous platforms employ LiDAR, radar, and ultrasonic sensors for metric depth perception. LiDAR provides centimeter-level point-cloud accuracy beyond 100~m but costs \$500-5,000 per automotive-grade unit, consumes 10-20~W, and creates significant packaging constraints that conflict with vehicle aesthetics and aerodynamics~\cite{wartnaby2023}. Radar at \$100-300 per unit offers lower angular resolution and struggles with stationary object discrimination. These burdens confine full active-sensor suites to premium vehicles, excluding the mass market from ADAS safety benefits.

Monocular cameras cost \$20 - 50 in production volumes, draw under 2~W, and integrate discreetly behind the windshield, making them the only sensor feasible for universal consumer-vehicle deployment. However, monocular images inherit a fundamental geometric deficiency: projecting a world point $(X,Y,Z)$ to image coordinates $(u,v) = (fX/Z,\,fY/Z)$ destroys metric depth, making recovery ill-posed without additional constraints~\cite{li2024acm}.

\subsection{Deep Learning for Monocular Depth}

Supervised deep networks have achieved remarkable qualitative progress on this problem. Eigen~\etal~\cite{eigen2014} introduced the first multi-scale convolutional depth regressor, demonstrating that networks can learn monocular depth cues from large training sets. Self-supervised approaches SfMLearner~\cite{zhou2017} and Monodepth2~\cite{godard2019} replaced expensive LiDAR supervision with photometric view-synthesis consistency. MiDaS~\cite{ranftl2020} achieved cross-domain transfer through a scale-and-shift-invariant loss trained on thirteen mixed datasets. DPT~\cite{ranftl2021} replaced the CNN encoder with a Vision Transformer for sharper depth boundaries, and Depth Anything~\cite{yang2024} scaled to unlabelled internet images via teacher-student distillation. ZoeDepth~\cite{bhat2023} addressed the metric scale problem by appending a metric head, comprising a seed-bin initializer and an attractor-based bin adjustment module, on top of a frozen MiDaS-DPT backbone. A domain-routing classifier selects between NYU-Depth-v2-trained and KITTI-trained metric heads at inference time, enabling zero-shot metric depth without per-scene calibration. Comprehensive surveys of these and related methods are provided by Masoumian~\etal~\cite{masoumian2022}, Li~\etal~\cite{li2024acm}, and Rajapaksha~\etal~\cite{rajapaksha2024}.

Despite these advances, deep monocular methods face three fundamental challenges for safety-critical deployment. First, they require large-scale LiDAR-supervised training data across diverse geographies, weather conditions, and lighting conditions. Second, domain shift causes substantial performance degradation when operational environments differ from training distributions: a model trained on clear Californian highways may fail catastrophically in Michigan winter conditions or night-time rain, as demonstrated by recent robustness evaluations of zero-shot depth models in natural outdoor environments~\cite{wildlife2025}. Third, network predictions lack explicit geometric reasoning, making it impossible to bound the error magnitude or identify the failure cause, a fundamental obstacle to certification under ISO~26262 Functional Safety~\cite{iso26262} and SOTIF (ISO/PAS~21448)~\cite{iso21448} standards.

\subsection{Geometric Priors and License Plate Ranging}

Geometric priors resolve scale ambiguity without training data and with fully interpretable physics-based reasoning. Lane-marking methods~\cite{stein2003} assume constant 3.7 m lane width but fail when markings are absent, faded by weather, or painted over in construction zones. Vehicle-dimension methods~\cite{choi2012} suffer from 400+ mm width variation across model classes. Ground-plane methods~\cite{han2016} require flat-road assumptions violated on hills, bridges, and driveways; even a $2^\circ$ pitch angle introduces errors exceeding 10\%. An end-to-end deep learning approach to inter-vehicle distance and velocity estimation was proposed by Song~\etal~\cite{song2020}, yet this requires extensive training data and domain-specific fine-tuning.

License plates present a uniquely advantageous geometric prior: they are government-mandated, dimensionally standardized, and attached to virtually every vehicle on the road. Standard U.S.\ plates measure $305\times152\,\text{mm}$ with FHWA-specified character heights varying from 63~mm (Tennessee, Texas) to 72~mm (Michigan), providing a known physical reference that resolves scale through elementary similar-triangle geometry, no training required. Furthermore, the combination of known geometry with physically interpretable measurements enables safety certification, since errors can be bounded analytically through error propagation.

Wang~\etal~\cite{wang2012} first demonstrated license plate-based ranging using plate width with 5\% relative error at 15~m. Rezaei~\etal~\cite{rezaei2014} added Kalman tracking for velocity estimation. Karagiannis~\cite{karagiannis2016} conducted thorough single-feature error analysis. Liu~\etal~\cite{liu2021} studied camera attitude compensation for inter-vehicle distance, motivating the pose compensation module in \tmdeE{}. MDE-Net~\cite{mdenet2022} proposed a neural network that fuses the keyhole imaging principle with object detection, but relies on training data. None of the geometric plate-based works applied state-specific character heights, compensated for camera pose, integrated deep-learning depth, or performed multi-character statistical averaging.

\subsection{Contributions}

Our prior work~\cite{reddy2026idetc} introduced \tmde{}: a geometric core demonstrating 2.3\% coefficient of variation in character height measurements and 36\% lower estimate variance than plate-width methods. The present paper extends this to \tmdeE{} with three contributions.

First, a four-method parallel plate detector (adaptive Gaussian, Otsu, Canny-dilation, bilateral-filter) with composite candidate scoring and automatic strict/permissive mode switching enables robust plate reading across the full range of automotive lighting and distance conditions. Second, a three-stage state identification engine fuses optical character recognition (OCR) text matching with 90+ plate markers, simultaneous multi-design HSV color scoring across all 127 cataloged designs, and an optional MobileNetV3-Small CNN~\cite{howard2019} classifier. Third, hybrid MiDaS DPT-Hybrid depth fusion via inverse-variance weighting and online EMA scale alignment, combined with a one-dimensional constant-velocity Kalman filter, provides smoothed distance, relative velocity, and TTC for FCW-compliant collision warning~\cite{nhsta_fcw}.

The paper is structured as follows: Section~\ref{sec:related_work} reviews related work; Section~\ref{sec:framework} describes the system architecture; Section~\ref{sec:implementation} details each algorithmic component; Section~\ref{sec:experimental} presents experimental validation; Section~\ref{sec:limitations} discusses limitations and failure modes; Section~\ref{sec:future} outlines future work; and Section~\ref{sec:conclusion} concludes.

\section{Related Work}\label{sec:related_work}

\subsection{Autonomous Vehicle Perception Architectures}

Autonomous vehicle perception must integrate multiple sensing modalities reliably and in real time. The Stanford \emph{Junior} vehicle winner of the 2007 DARPA Urban Challenge combined GPS/IMU/LiDAR localization achieving ${<}10$~cm lateral RMS accuracy with boosted-classifier object recognition and Kalman-filter tracking~\cite{levinson2011}. A recent comprehensive survey of 3D object detection for autonomous driving~\cite{contreras2024} identifies camera-based monocular methods as the most cost-effective pathway to mass-market deployment, while multi-modal fusion remains the gold standard for performance. \tmdeE{} is designed to slot into the camera-centric perception layer of these architectures, providing a certifiable, interpretable distance estimate for a lead vehicle without additional active sensors.

\subsection{Classical Geometric Monocular Distance Estimation}

Classical monocular ranging resolves scale ambiguity by leveraging known physical dimensions of scene elements as geometric constraints. Stein~\etal~\cite{stein2003} developed the first vision-based adaptive cruise control system using the known 3.7m U.S.\ lane width as a distance reference, additionally estimating camera pitch from the vanishing point of lane lines. While effective on well-marked highways, this fails in construction zones, parking facilities, and unmarked rural roads. Choi~\etal~\cite{choi2012} used Haar-like features and vehicle bounding-box width; however, vehicle width varies by more than 400~mm between a compact car and a full-size SUV. Han~\etal~\cite{han2016} exploited vehicle contact-point geometry with flat-road assumptions; even a $2^\circ$ pitch angle introduces errors exceeding 10\%. Liu~\etal~\cite{liu2021} explicitly modelled camera attitude angles and proposed a compensation technique for inter-vehicle distance estimation, confirming the quantitative impact of pose on ranging accuracy.

\subsection{Deep Learning-Based Monocular Depth}

The application of deep convolutional networks to depth estimation transformed the field. Eigen~\etal~\cite{eigen2014} first demonstrated that a multi-scale network could predict dense depth maps from single images by learning monocular depth cues. Self-supervised methods SfMLearner~\cite{zhou2017} and Monodepth2~\cite{godard2019} removed the need for LiDAR supervision. MiDaS~\cite{ranftl2020} trained on thirteen mixed datasets with a scale-and-shift-invariant loss for zero-shot cross-dataset transfer. DPT~\cite{ranftl2021} replaced the CNN backbone with a Vision Transformer. ZoeDepth~\cite{bhat2023} addressed metric scale through the same seed-bin/attractor architecture described above, achieving a 21\% relative improvement in absolute relative error on NYU-Depth v2 over the prior state of the art. Depth Anything~\cite{yang2024} achieved state-of-the-art zero-shot generalization via teacher-student distillation on unlabeled internet images. Comprehensive surveys are available in Masoumian~\etal~\cite{masoumian2022}, Li~\etal~\cite{li2024acm}, and Rajapaksha~\etal~\cite{rajapaksha2024}. ZoeDepth has been shown to degrade significantly in natural outdoor environments~\cite{wildlife2025}, confirming that domain shift remains unsolved.

Song~\etal~\cite{song2020} proposed an end-to-end deep network for inter-vehicle distance and velocity estimation from monocular video, integrating optical flow and scene geometry, but requiring domain-specific training data. Despite impressive results, all deep methods share the fundamental limitation that their predictions cannot be bounded analytically for safety certification purposes, unlike the geometric prior approach of \tmdeE{}.

\subsection{License Plate-Based Distance Estimation}

License plates provide a uniquely reliable geometric prior~\cite{wang2012, rezaei2014,karagiannis2016}. Wang~\etal~\cite{wang2012} achieved 5\% relative error using plate width at ranges up to 15~m. Rezaei~\etal~\cite{rezaei2014} integrated Kalman tracking, showing that the differentiated pinhole equation $\dot{D} = -D\dot{h}/h$ provides relative velocity from the rate of change of apparent plate height, eliminating the need for any additional sensor for velocity estimation. Karagiannis~\cite{karagiannis2016} conducted the most thorough single-feature error analysis, modeling contributions from camera height, road gradient, plate rotation, and calibration uncertainty.

MDE-Net~\cite{mdenet2022} proposed a neural network fusing the keyhole imaging principle with YOLOv5 plate detection, reducing error to below 3\% in urban scenarios. A monocular vision study using number plate parameters~\cite{nvplate2024} employed regression and ANN models on plate height/width features. None of these works applied state-specific FHWA character heights, compensated for camera pose dynamics, integrated complementary deep-learning depth, or performed multi-character statistical averaging with outlier rejection. \tmdeE{} directly addresses all four gaps.

\subsection{Automatic License Plate Recognition}

The ALPR field has produced highly relevant detection and OCR techniques. Laroca~\etal~\cite{laroca2018} demonstrated the first YOLO-based end-to-end ALPR system. Silva \& Jung~\cite{silva2022} extended this to a three-stage pipeline. Zhao~\etal~\cite{yolov5lprnet} improved adverse-weather robustness with attention-augmented YOLOv5s and Soft-NMS. A comprehensive review of deep learning for ALPR~\cite{dlalpr2026} surveys YOLO variants, CRNN-based recognizers, and transformer architectures. Satya~\etal~\cite{yolov8alpr2025} demonstrated an optimized YOLOv8 ALPR pipeline for resource-constrained devices using TensorRT INT8 quantization, achieving real-time throughput on embedded hardware.

\tmdeE{} fundamentally differs from ALPR: while ALPR reads the plate serial number for identification, \tmdeE{} uses the physical dimensions of characters (height, stroke width, inter-character spacing) as geometric references for metric distance estimation. The OCR subsystem in \tmdeE{} serves for state identification and serial format validation, not vehicle identification.

\subsection{Camera Pose Compensation}

Camera pitch and roll during vehicle dynamics, acceleration, braking, cornering, road undulations, introduce systematic errors in geometric ranging. Liu~\etal~\cite{liu2021} studied these effects comprehensively, showing errors beyond 10\% even for moderate pitch angles. Stein~\etal~\cite{stein2003} estimated pitch from the vanishing point of lane lines. Workman~\etal~\cite{workman2016} trained a CNN to predict horizon lines without visible lane markings. The \tmdeE{} pose compensation module implements a hybrid approach: a geometric estimation based on the Probabilistic Hough Transform when lane markings are visible, with the deep horizon predictor as a fallback.

\subsection{Temporal Filtering and Collision Warning}

Kalman filtering for plate-based tracking and velocity estimation was pioneered by Rezaei~\etal~\cite{rezaei2014} and extended to multi-sensor fusion by Moon~\etal~\cite{moon2016}. A survey of autonomous-vehicle collision avoidance algorithms~\cite{cai2025collision} identifies Kalman-based TTC estimation as the dominant monocular camera approach. NHTSA FCW activation requires TTC $\leq 2.4$~s~\cite{nhsta_fcw}.

\subsection{Efficient CNN Architectures for Embedded ADAS}

MobileNetV3-Small~\cite{howard2019} is used as the CNN state classifier in the third stage. Its hardware-aware NAS design achieves 6.6\% higher ImageNet accuracy than MobileNetV2-Small at similar latency, with 2.5~M parameters fitting within the GPU memory budget alongside MiDaS. Howard~\etal~\cite{howard2019} confirm that MobileNet-family architectures are well-suited for real-time vision tasks on edge ADAS hardware.

\section{T-MDE Enhanced System Architecture}\label{sec:framework}

\begin{figure}[t]
\centering
\includegraphics[width=0.8\textwidth]{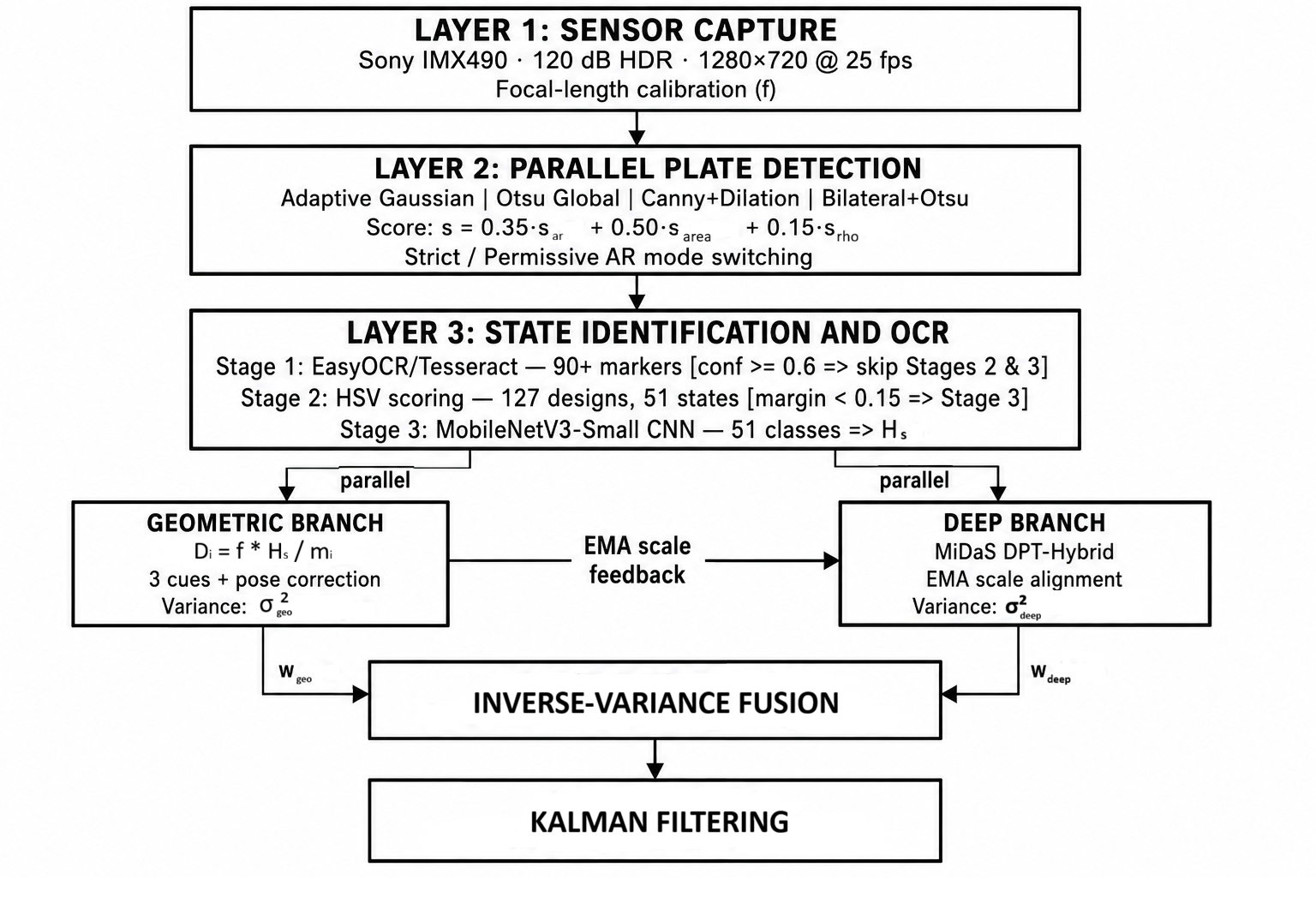}
\caption{\tmdeE{} four-layer system architecture.}
\label{fig:architecture}
\end{figure}

The \tmdeE{} system is organized into four functional layers, with solid arrows in Figure~\ref{fig:architecture} denoting primary data flow: Sensor $\rightarrow$ Detection $\rightarrow$ State ID and OCR $\rightarrow$ Distance Fusion and Warning.

Figure~\ref{fig:architecture} illustrates the four-layer pipeline. Layers~1 - 3 are strictly sequential: Layer~2 runs four binarization methods in parallel and converges on a single plate candidate via composite scoring; Layer~3 implements a sequential short-circuit (Stage~1 $\rightarrow$ 2 $\rightarrow$ 3) so that a confident OCR match bypasses color and CNN stages entirely; Layer~4 runs the geometric and deep branches in parallel before inverse-variance fusion and Kalman filtering.

\textbf{Layer 1: Sensor Capture.}
The camera capture layer interfaces with the automotive-grade camera (Sony IMX490 sensor with GW5400 ISP) via standard USB Video Class (UVC) drivers. The camera provides 120~dB HDR tone mapping, which is critical for handling the wide dynamic range of outdoor driving scenes (bright sky, dark vehicle underside). The capture resolution can be configured; for primary evaluation we use $1280\times720$ to balance detail and frame rate. The camera's focal length is known from calibration (Section~\ref{sec:calibration}) and is used in all subsequent geometric calculations.

\textbf{Layer 2: Plate Detection.}
The plate detection layer runs four different preprocessing and binarization methods in parallel on every incoming frame: adaptive Gaussian thresholding (handles uneven illumination), Otsu's method (optimal for high-contrast outdoor scenes), Canny edge detection followed by morphological dilation (effective in dim light or near-infrared conditions), and bilateral filtering combined with Otsu (reduces sensor noise and motion blur). Each method produces a set of candidate contours. A composite scoring function, combining aspect ratio, area, and edge density, selects the most likely plate region. The system automatically switches between strict and permissive aspect ratio bounds after eight consecutive missed detections, allowing it to adapt to challenging conditions without user intervention.

\textbf{Layer 3: State Identification and OCR.}
Once a plate is detected, the perspective distortion is removed via a homography that maps the four detected corners to a fronto-parallel rectangle of known dimensions ($305\times152$~mm). The rectified plate image is then processed by the three-stage state identification engine. Stage 1 extracts text from the top and bottom strips (where state names and slogans appear) using an OCR engine. Stage 2 analyzes the color distribution in HSV space and compares it against a pre-cataloged set of 127 plate designs for all 51 jurisdictions. Stage 3 passes the rectified plate image to a lightweight convolutional neural network (MobileNetV3-Small) that outputs a probability distribution over states. The identified state determines the FHWA character height $H_s$ used in distance calculation.

\textbf{Layer 4: Distance Estimation and Sensor Fusion.}
The distance estimation layer operates three parallel branches. The geometric branch computes distance using the pinhole camera model: $D = f H_s / \bar{h}$, where $\bar{h}$ is the outlier-robust mean character height obtained from the character segmentation module. The pose compensation branch corrects $\bar{h}$ for camera pitch and roll estimated from lane markings (or a deep horizon predictor). The deep branch runs the MiDaS DPT-Hybrid network on the full frame to obtain a relative depth map, then aligns its scale online to the geometric estimate using an exponential moving average. The two metric estimates ($D_{\mathrm{geo}}$ and $D_{\mathrm{deep}}$) are fused via inverse-variance weighting. A one-dimensional constant-velocity Kalman filter smooths the fused distance and provides relative velocity and TTC.
Table~\ref{tab:hardware} summarizes the development and validation hardware.

\begin{table}[htbp]
\caption{Development and validation of hardware.}
\label{tab:hardware}
\centering\small
\begin{tabular}{ll}
\toprule
\textbf{Component} & \textbf{Specification} \\
\midrule
Camera sensor  & Sony IMX490, 1/1.55" CMOS, 3.0~$\mu$m pixel pitch \\
ISP              & GEO Semiconductor GW5400, 120~dB HDR \\
Serializer       & Maxim MAX9295A/B GMSL2 \\
Interface        & LI-GMSL2-FP-USB-BOX $\rightarrow$ USB 3.0 (UVC) \\
Full resolution  & $2880\times1860$ at 25~fps \\
Working res.     & $1280\times720$ (primary evaluation) \\
Primary lens     & 040H: EFL 11.9~mm, FOV H 38$^\circ$, IP67 \\
GPU              & NVIDIA RTX 4070 (Acer Predator Helios 16) \\
CUDA / PyTorch   & 12.4 / 2.6.0+cu124 \\
CPU / OS         & Intel Core i7 13th Gen, Windows 11 \\
\bottomrule
\end{tabular}
\end{table}

\subsection{System Operation Flow and Decision Logic}\label{sec:flow}

To clarify the distinction between sequential and parallel processing
stages, Figure~\ref{fig:architecture} and the text below describe
the precise control flow on each video frame.

\textbf{Per-frame sequential steps.}
(1)~The camera delivers a raw frame (Layer~1).
(2)~All four binarization methods execute \emph{in parallel}; their candidate contours are merged and scored; a single plate bounding box is either confirmed or the frame is flagged as a missed detection (Layer~2, parallel $\rightarrow$ sequential convergence).
(3)~The confirmed plate crop is rectified via homography. The three-stage state identification engine then proceeds \emph{sequentially}: Stage~1 (OCR) is run first and, if it yields confidence~$\geq 0.6$, Stages~2 and~3 are skipped entirely.  If Stage~1 is ambiguous, Stage~2 (HSV) runs; if Stage~2 margin is $< 0.15$, Stage~3 (CNN) is invoked (Layer~3, sequential short-circuit logic).
(4)~Given the confirmed $H_s$, two branches run \emph{in parallel}: (a)~the geometric branch computes $D_{\mathrm{geo}}$ from three typographic cues with outlier-robust averaging and pose correction; (b)~the MiDaS branch extracts $D_{\mathrm{deep}}$ via EMA scale alignment.  These are fused immediately afterwards via inverse-variance weighting (Layer~4, parallel $\rightarrow$ sequential convergence).
(5)~The Kalman filter receives $D_{\mathrm{fused}}$ (or executes a predict-only step if no plate was detected) and outputs $\hat{D}_k$, $\hat{v}_k$, and TTC.

\textbf{Override / fallback logic.}
Table~\ref{tab:decision_logic} summarises the five operating modes and the output each produces, making explicit how intermediate results are fused or overridden at runtime.

\begin{table}[htbp]
\centering\small
\caption{Per-frame operating modes and distance-output sources.}
\label{tab:decision_logic}
\begin{tabular}{llll}
\toprule
\textbf{Mode} & \textbf{Plate?} & \textbf{MiDaS?} & \textbf{Distance source} \\
\midrule
Normal        & Yes  & Yes  & Inv.-var.\ fused $D_{\mathrm{fused}}$ \\
Geo-only      & Yes  & No   & $D_{\mathrm{geo}}$ (geometric branch) \\
Deep-only     & No   & Yes  & $s_t \cdot d_{\mathrm{plate}}$ (held $s_t$) \\
Predict-only  & No   & No   & Kalman constant-vel.\ prediction \\
State fallback& Yes  & Yes  & $D_{\mathrm{fused}}$ with $H_s = 65.1$~mm \\
\bottomrule
\end{tabular}
\end{table}

The \emph{predict-only} mode is bounded: after more than 25 consecutive prediction-only frames (about 1~s at 25~fps), a confidence flag is raised and TTC warnings are suppressed until at least one fused or deep-only measurement is received.

\section{Implementation Details}\label{sec:implementation}

\subsection{Camera Calibration}\label{sec:calibration}

Accurate focal length is essential for metric distance estimation. The system performs an interactive calibration using a license plate as a known reference. The key innovation is using segmented character height not the overall plate warp height for calibration. Perspective warp inflation can cause the plate bounding box to appear over 50\% taller than its true pixel height, producing physically implausible focal lengths (e.g., $f = 12{,}224$~px). Character height from the segmentation algorithm (described below) is immune to such inflation because its geometric filters reject oversize regions.

\begin{table}[h]
\caption{Lens variants with datasheet-derived focal lengths.}
\label{tab:focal_lengths}
\centering\small
\begin{tabular}{lcccc}
\toprule
\textbf{Lens} & \textbf{EFL (mm)} & \textbf{FOV H}
            & \textbf{$f$ at 2880~px} & \textbf{$f$ at 1280~px} \\
\midrule
030H (CAR-56) & 16.37 & 30$^\circ$ & 5,457~px & 2,425~px \\
040H (DSL612) & 11.90 & 38$^\circ$ & 3,967~px & 1,763~px \\
065H (DSL608) &  7.90 & 65$^\circ$ & 2,633~px & 1,170~px \\
120H (CAR-55) &  4.49 & 120$^\circ$ & 1,497~px & 665~px \\
\bottomrule
\end{tabular}
\end{table}

The user places a known plate (with known $H_s$) at three distances $D_{\mathrm{ref}} = 1.0, 2.0, 3.0$~m. At each distance, the system detects the plate, extracts characters, and computes the mean character height $\bar{h}$. A candidate focal length is then computed as $f = \bar{h} \cdot D_{\mathrm{ref}} / H_s$. The three candidates are aggregated using the median, which is robust to a single outlier measurement. The resulting focal length is stored and scales linearly to any capture resolution: $f_{\mathrm{cap}} = f_{\mathrm{cal}} \cdot W_{\mathrm{cap}} / W_{\mathrm{cal}}$.
Table~\ref{tab:focal_lengths} lists the four lens variants available for the LI-IMX490 camera, with their corresponding focal lengths at full resolution ($2880\times1860$) and at the working resolution ($1280\times720$).

\subsection{Four-Method Parallel Plate Detection}\label{sec:plate_detection}

\begin{figure}[ht]
\centering
\includegraphics[
    width=0.7\linewidth,
]{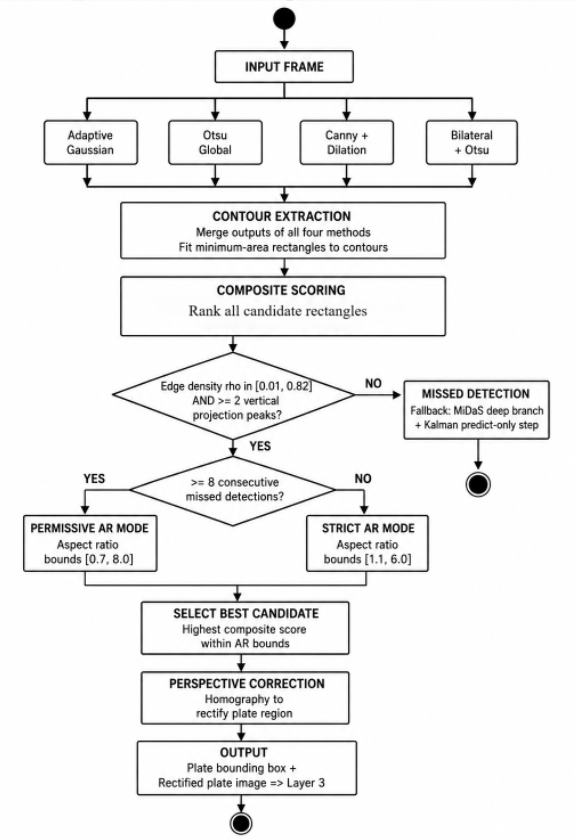}
\caption{Parallel plate detection and candidate selection flowchart.}
\label{fig:flowchart_detection}
\end{figure}

Figure~\ref{fig:flowchart_detection} shows the complete detection pipeline. Four parallel preprocessing branches run simultaneously on every frame; a composite score selects the best plate candidate, followed by a geometric verification step and adaptive mode switching between strict and permissive aspect ratio bounds. The plate detection module processes every incoming video frame using four parallel pipelines, each applying a distinct combination of preprocessing and thresholding techniques, handling a wide range of illumination conditions.

\textbf{Adaptive Gaussian thresholding.} The first pipeline applies an adaptive Gaussian threshold that computes a local threshold for each pixel based on the weighted mean of its neighborhood. A window size of \(11\times11\) pixels is used, and a constant is subtracted from the mean to separate foreground (plate characters and borders) from background. This handles uneven illumination where a global threshold would fail.

\textbf{Otsu's method.} The second pipeline employs Otsu's global thresholding algorithm, which finds a single intensity threshold that minimizes the intra‑class variance (i.e., maximizes the inter‑class variance) between foreground and background pixels. This works well for high-contrast outdoor scenes under daylight.

\textbf{Canny edge detection followed by morphological dilation.} The third pipeline first detects edges using the Canny operator with hysteresis thresholds set to 30 (low) and 100 (high). This extracts strong and weak edges while suppressing noise. The resulting binary edge map is then dilated using a \(5\times5\) square kernel, which closes small gaps in the plate contours caused by broken edges or low contrast. This is robust in dim light or near-infrared conditions.

\textbf{Bilateral filtering combined with Otsu.} The fourth pipeline applies a bilateral filter (diameter 9 pixels, sigma 75) to the input image before thresholding. The bilateral filter reduces sensor noise and smooths flat regions while preserving sharp edges, which is crucial for motion‑blurred or low‑quality images. After filtering, Otsu's method is applied to obtain a binary image. This combination handles rainy conditions, camera shake, and compression artifacts.

\begin{figure}[htbp]
\centering
\includegraphics[width=0.5\linewidth]{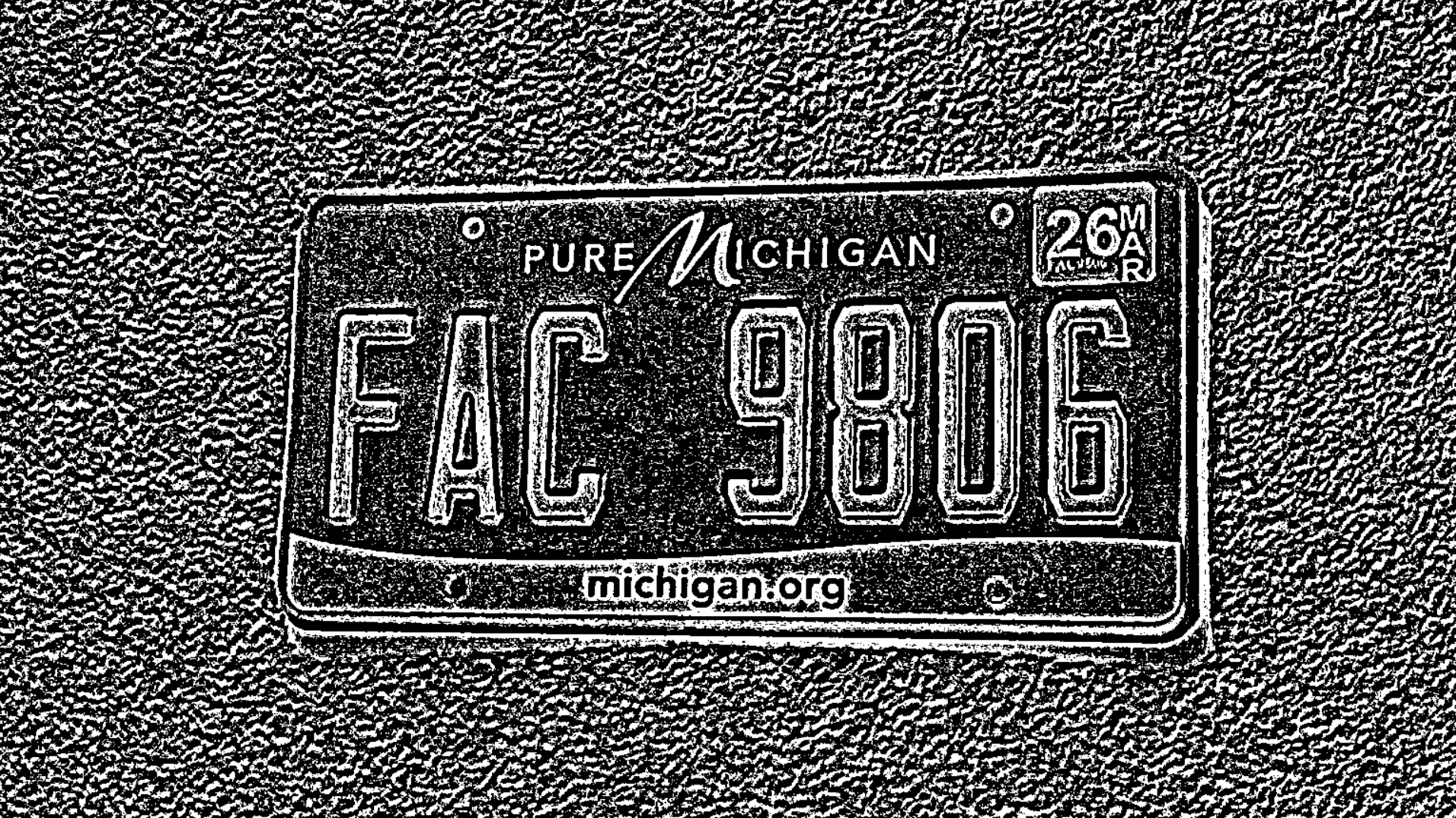}
\caption{Adaptive Gaussian binarisation output for Michigan license plate.}
\label{fig:adaptive_binary}
\end{figure}

Figure~\ref{fig:adaptive_binary} shows a representative output of the adaptive Gaussian thresholding pipeline on a Michigan license plate (FAC~9806). The binarised image clearly resolves the character strokes against the plate background, with the plate border and character regions emerging as distinct contours despite the textured wall surface behind the plate. This confirms that the local-window threshold effectively separates plate foreground from uneven ambient illumination.

After all four pipelines have produced their sets of contour candidates, the system processes each candidate by fitting a minimum‑area rectangle around it and computing three geometric properties: the aspect ratio \(ar = w/h\) (width divided by height), the area of the rectangle relative to the total image size, and the edge density \(\rho\) inside the candidate region (the ratio of edge pixels to total pixels). For a genuine license plate, the edge density typically lies around \(\rho \approx 0.12\). A composite score \(s = 0.35\,s_{ar} + 0.50\,s_{\mathrm{area}} + 0.15\,s_\rho\) is then calculated, where each component \(s_{\cdot}\) is a normalized measure. For example, the aspect ratio score is defined as \(s_{ar} = 1 - |ar - 2.5| / 2.5\), clipped to the interval \([0,1]\), reflecting the ideal aspect ratio of a US license plate (approximately 2.5). The candidate with the highest composite score is selected as the plate candidate, provided it also passes a verification step: the edge density must lie within \([0.01, 0.82]\) and the vertical projection of the candidate region must contain at least two distinct peaks (indicating the presence of multiple characters).

To handle extreme viewing angles where the plate appears severely skewed (e.g., during sharp turns or when the lead vehicle is very close), the system implements an adaptive mode‑switching mechanism. Under normal operation, the system uses a strict aspect ratio range of \(ar \in [1.1, 6.0]\). However, after eight consecutive frames with no successful detection, the system automatically switches to a permissive mode where the aspect ratio range is expanded to \(ar \in [0.7, 8.0]\), allowing it to catch highly distorted plates. As soon as a detection succeeds, the system reverts to strict mode for the next frame. This adaptive behavior ensures high precision in typical driving scenarios while maintaining robustness during brief adverse conditions.

\subsection{Dual-Threshold Character Segmentation}\label{sec:char_seg}

Once a license plate has been successfully detected and its perspective distortion has been removed via a homography that maps the four detected corners to a fronto‑parallel rectangle of known dimensions (\(305\times152\) mm), the character segmentation module takes the rectified plate image and extracts individual bounding boxes for each character. The height of the rectified plate image varies significantly with distance: at close range (e.g., 3 - 5m), the plate height is typically 150 - 200 pixels, providing ample resolution for accurate segmentation. However, at longer distances (e.g., 20~m), the plate height may shrink to only 30 - 40 pixels, making character extraction challenging. To address this, the module first checks whether the plate height \(h_p\) is less than 100 pixels. If so, it resizes the plate by a factor \(s = \max(2.0, 100/h_p)\) using bicubic interpolation. This rescaling ensures that the plate height is at least 100 pixels, improving the visibility of small characters while preserving edge sharpness (bicubic interpolation is preferred over bilinear because it produces smoother edges and reduces aliasing artifacts). The factor is capped at a minimum of 2.0 to avoid excessive scalings that could introduce noise or blur.

The segmentation algorithm applies two complementary thresholding methods in parallel: adaptive Gaussian thresholding and Otsu's global thresholding. Both methods run simultaneously on the (possibly resized) rectified plate image. The adaptive Gaussian method uses a local window to compute a threshold for each pixel, which is effective for plates with uneven illumination (e.g., shadows falling across the plate). Otsu's method computes a single global threshold that minimizes the intra‑class variance, which works well when the plate has uniform lighting and high contrast. For each of the two resulting binary images, the module applies morphological opening (erosion followed by dilation) and closing (dilation followed by erosion) using a \(3\times3\) square kernel. Opening removes small noise pixels and separates lightly connected characters, while closing fills small gaps and holes within character strokes. After morphological processing, contours are extracted from each binary image.

Each extracted contour is then evaluated with five geometric criteria designed to accept only valid character regions and reject non‑character artifacts such as plate borders, state name strips, reflections, or merged characters. First, the contour height must be between \(0.2h_p\) and \(0.8h_p\); this excludes the top and bottom strips of the plate (which typically contain the state name, slogan, or decorative graphics) and also rejects the plate's outer border. Second, the contour width must be less than \(1.8\) times its height; this rejects cases where two or more characters have merged together due to insufficient gap or poor binarization. Third, the aspect ratio \(w/h\) must lie between \(0.15\) and \(1.5\); this range covers the narrowest character (the digit "1" or the letter "I") through the widest character (the letter "W" or "M"). Fourth, the minimum dimensions are set to \(h \ge 5\) pixels and \(w \ge 2\) pixels, which rejects tiny noise blobs that are too small to be genuine characters. Fifth, the vertical position of the contour must lie within the central 80\% of the plate height, which excludes reflections or specular highlights that often appear near the top or bottom edges of the plate frame.

After applying these five filters independently to the contours from both thresholding methods, the module performs a \(2\sigma\) outlier rejection on the set of surviving character heights for each method. Specifically, it computes the mean \(\mu_h\) and standard deviation \(\sigma_h\) of the heights, then removes any character whose height deviates from the mean by more than \(2\sigma_h\). This step eliminates any remaining anomalous characters that passed the geometric filters (e.g., a partially occluded character or a decorative graphic that accidentally resembles a character shape). The method (adaptive Gaussian or Otsu) that yields the larger number of valid characters after outlier rejection is selected as the winner for that frame. The final mean character height \(\bar{h}\) is then computed as the arithmetic mean of the heights of the surviving characters from the chosen method:

\begin{equation}
\bar{h} = \frac{1}{n}\sum_{i=1}^{n} h_i
\label{eq:avg_height}
\end{equation}
where \(n\) is the number of valid characters (typically between 3 and 8 for a US license plate). If after outlier rejection fewer than three characters survive, the segmentation is considered a failure; the frame is skipped and no distance estimate is produced from the geometric branch for that frame. This threshold of three characters ensures that the mean height estimate is statistically meaningful and reduces the risk of large errors from a single mis‑segmented character. In such failure cases, the system either relies on the deep learning branch (MiDaS) if available, or uses the Kalman filter to predict the distance from previous frames until a valid segmentation is again possible.

\subsection{Three-Stage State Identification}\label{sec:state_id}

The state identification module determines which U.S. state (or the District of Columbia) issued the detected license plate. This determines the FHWA character height \(H_s\), and using the wrong height introduces systematic ranging errors (up to 10.6\% for Michigan plates). The module operates in three sequential stages; later stages are invoked only when earlier stages produce ambiguous results.

\textbf{Stage 1: Text OCR.}
The top 28\% and bottom 22\% of the rectified plate image are cropped to isolate the regions where state names and slogans typically appear. These crops are passed to an OCR engine running on the GPU, specifically EasyOCR, which combines a CRAFT (Character Region Awareness for Text Detection) detector for text localization with a CRNN (Convolutional Recurrent Neural Network) recognizer for character transcription. The recognized text strings are then matched against a dictionary of over 90 pre-stored textual markers, which include full state names (e.g., "MICHIGAN"), official slogans (e.g., "PURE MICHIGAN", "EMPIRE STATE", "LONE STAR STATE"), website addresses (e.g., "michigan.org"), and common phrases like "In God We Trust" or "Taxation Without Representation". Matches are prioritized by two criteria: first by the length of the matched phrase (longer phrases are more specific and thus more reliable), and second by the OCR confidence score. For example, a detection of "PURE MICHIGAN" with confidence 0.92 takes priority over a detection of "MICHIGAN" alone with confidence 0.80, because the full slogan is unique to Michigan whereas the word "MICHIGAN" could appear on other plates (e.g., "MICHIGAN GREAT LAKES"). Partial matches (e.g., "MICH" when the rest is occluded) are considered only if no full match is found. If EasyOCR returns an empty result (e.g., due to severe blur, dirt, or non-standard fonts), the system falls back to Tesseract 5 with page segmentation mode 7 (treating the image as a single text line) to provide a secondary OCR attempt. Stage 1 is given infinite weight in the final decision: if it returns a match with confidence above a threshold (typically 0.6), that match immediately determines the state, and Stages 2 and 3 are bypassed entirely. This design choice reflects the fact that textual markers, especially state names and unique slogans, are the most reliable identifiers, far more discriminative than color or even deep learning features.

\textbf{Stage 2: Multi-Design HSV Scoring.}
When Stage 1 does not produce a confident match (e.g., the plate has no text or the text is illegible), the system proceeds to Stage 2, which scores all 127 plate designs across all 51 jurisdictions simultaneously against the color distribution of the rectified plate. For each design \(j\) belonging to state \(s\), the module counts the number of pixels in the plate image whose HSV (hue, saturation, value) values fall within the design's pre-stored HSV range(s). Multiple ranges per design are stored to account for lighting variations, shadows, and specular highlights. The score for state \(s\) is computed as:

\begin{equation}
\mathrm{score}_s = \sum_{j \in \mathcal{D}_s} w_j \cdot
  \frac{N_{\mathrm{roi}}^{(j)}}{N_{\mathrm{roi}}}
\label{eq:hsv_score}
\end{equation}
where \(N_{\mathrm{roi}}\) is the total number of pixels in the rectified plate image, \(N_{\mathrm{roi}}^{(j)}\) is the number of pixels falling within the HSV range(s) of design \(j\), and \(w_j\) is a design-specific weight that is higher for common designs (e.g., standard-issue plates) and lower for rare specialty or vintage plates. This weighted sum ensures that a plate matching a common design receives a higher score than one matching only a rare design, even if the raw pixel count is the same. An important feature is that all 127 designs are scored simultaneously, not pairwise. This prevents a Michigan Mackinac Bridge plate (gold/orange sunset background) from being misidentified as Florida (which also has orange accents) because the combined score from Michigan's three active designs (standard, Mackinac Bridge, and any other variant) will always exceed Florida's score. Six states with uniquely colored backgrounds receive a confidence boost of \(+0.20\) when a color match is detected. These states are: Delaware (black background with gold text the only US plate with a black background), New Jersey (uniform yellow-to-white gradient), Vermont (solid green background), Alaska (gold background with Big Dipper stars), Oklahoma (bold red background with Star-46 icon, introduced in September 2024), and New Mexico (yellow background with red sun symbol). The boost is applied directly to the state's score before normalization, giving these visually distinctive states an advantage when their color signature is present. After scoring, the scores are normalized to sum to 1 across all states, and the state with the highest normalized score is tentatively selected provided that the margin above the second-best state exceeds a threshold (typically 0.15). If the margin is smaller, the system flags a confusion pair and invokes the rule-based disambiguation rules described in Section~\ref{sec:state_id}.

\textbf{Stage 3: CNN Classifier.}
Stage 3 is a deep learning fallback that is activated only when Stages 1 and 2 both produce confidences below 0.7, or when a specific confusion pair (e.g., North Carolina vs. Ohio) has been detected and the rule-based disambiguation did not yield a decisive winner. The classifier uses MobileNetV3-Small~\cite{howard2019}, a hardware-aware NAS-designed network built around inverted residual blocks with hard-swish activations and a Squeeze-and-Excitation attention module in each bottleneck. It contains 2.5~M parameters across 14 bottleneck stages and achieves 67.4\% top-1 accuracy on ImageNet, providing an excellent accuracy-to-latency trade-off for the 51-class state identification task. The network takes as input the rectified plate crop resized to \(128 \times 256\) pixels (preserving the 1:2 aspect ratio of US plates) and outputs a softmax vector over 51 classes (50 states plus the District of Columbia). When Stage~3 is activated, the system saves detected plate crops and their identified state labels to disk, allowing the user to build a local training set over time. The training protocol involves freezing the backbone for the first five epochs with a learning rate of \(10^{-3}\), followed by fine-tuning the full model for 45 epochs using cosine decay with an initial learning rate of \(10^{-4}\). Data augmentation includes random rotations of \(\pm10^\circ\), brightness adjustments of \(\pm0.3\), color jitter, and random erasing to improve generalization to real-world variations. When Stage 3 is activated, its softmax probabilities are combined with the normalized scores from Stage 2 using a weighted average (with Stage 2 weight typically 0.6 and Stage 3 weight 0.4), because Stage 2's color information remains valuable even when ambiguous. The final state decision is the one with the highest combined score. Once a state is identified, the corresponding character height \(H_s\) is looked up from the state table and used in all subsequent distance calculations. If all three stages fail to produce a confident decision (combined score below 0.5), the system defaults to a conservative \(H_s = 65.1\) mm (the national average) and logs the frame for later manual review and potential addition to the training set.

\subsection{Camera Pose Compensation Using Perspective Geometry}\label{sec:pose_compensation_detailed}

\begin{figure}[htbp]
\centering
\includegraphics[width=0.3\linewidth]{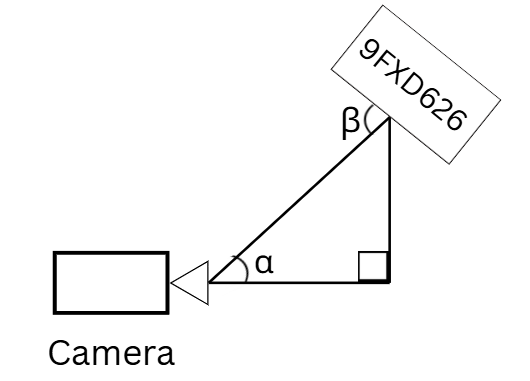}
\caption{Pose compensation geometry with pitch angle and the plate tilt angle.}
\label{fig:pose_compensation}
\end{figure}

Accurate distance estimation using the pinhole camera model requires knowledge of the camera's orientation relative to the target plate. When the camera is pitched upward or downward, the apparent height of the license plate characters is foreshortened, introducing systematic error into the distance calculation. Figure~\ref{fig:pose_compensation} illustrates this geometric relationship.

\textbf{Geometric interpretation.}
Figure~\ref{fig:pose_compensation} shows a camera mounted in a vehicle, observing a license plate attached to a lead vehicle. The camera has a pitch angle \(\phi\) relative to the horizontal plane: positive pitch means the camera is tilted upward (looking toward the sky), negative pitch means downward (looking toward the road). The true physical height of the license plate characters is \(H_s\) (a known state-specific value, e.g., 72 mm for Michigan). The distance along the optical axis from the camera to the plate is \(D\). Due to the pitch angle, the effective viewing angle causes the projected image height \(\bar{h}\) to be smaller than it would be if the camera were perfectly aligned (i.e., \(\phi = 0\)). The diagram also indicates the focal length \(f\) of the camera lens, which relates the physical geometry to the image plane coordinates. The foreshortening factor is \(\cos\phi\): when \(\phi = 0\), \(\cos\phi = 1\) and no correction is needed; as \(\phi\) increases, \(\cos\phi\) decreases, reducing the measured height. The figure also implies the presence of a roll angle \(\psi\) (rotation around the optical axis), which further compresses the projected height by a factor \(\cos\psi\). Together, these effects must be compensated to recover the true distance.

\textbf{Derivation of the basic perspective equation.}
The fundamental pinhole projection equation for a vertical object of true height \(H_s\) placed at distance \(D\) from a camera with focal length \(f\), when the camera is pitched by an angle \(\phi\) relative to the object plane, is:

\begin{equation}
\frac{\bar{h}}{f} = \frac{H_s}{D} \cdot \cos\phi
\label{eq:basic_perspective}
\end{equation}

This equation is derived from similar triangles in the perspective projection, accounting for the fact that the effective height seen by the camera is the projection of \(H_s\) onto a plane perpendicular to the optical axis. The term \(\cos\phi\) arises because the plate is tilted relative to the image plane. If \(\phi = 0\) (camera perfectly horizontal), the equation reduces to the standard pinhole formula \(\bar{h}/f = H_s/D\). For a non-zero pitch, the measured image height is reduced proportionally to \(\cos\phi\), which would cause the naive distance estimate \(D = f H_s / \bar{h}\) to overestimate the true distance (since a smaller \(\bar{h}\) implies a larger \(D\)). Therefore, we must correct for this foreshortening.

\textbf{Pose correction for pitch and roll.}
For a camera that is both pitched by \(\phi\) and rolled by \(\psi\) (rotation around the optical axis), the combined foreshortening effect multiplies the individual cosine factors. The corrected character height \(h'\) that would have been measured if the camera were perfectly aligned (\(\phi = 0, \psi = 0\)) is:

\begin{equation}
h' = \bar{h} \cdot \frac{1}{\cos\phi \cdot \cos\psi}
\label{eq:pose_correction_detailed}
\end{equation}

This equation is obtained by inverting the foreshortening: the true projected height equals the measured height divided by the product of the cosines. Substituting this corrected height into the standard pinhole equation gives the pose-compensated distance estimate:

\begin{equation}
D_{\mathrm{compensated}} = \frac{f \cdot H_s}{h'} =
  \frac{f \cdot H_s}{\bar{h}} \cdot \cos\phi \cdot \cos\psi
\label{eq:distance_with_pose}
\end{equation}

Observe that the compensation factor is exactly \(\cos\phi \cdot \cos\psi\), which multiplies the naive distance estimate \(f H_s / \bar{h}\). If both angles are zero, the factor is 1 and no compensation is applied. For small positive pitch (camera tilted upward), \(\cos\phi < 1\), so the compensated distance is smaller than the naive estimate by correcting the overestimation.

\textbf{Estimating pitch from lane markings.}
The pitch angle \(\phi\) is estimated from the vanishing point of lane markings. The Probabilistic Hough Transform detects lane lines in the image; their intersection (or the point where parallel lines appear to converge) gives the vanishing point. Given the image height \(H_I\) (in pixels) and the vertical coordinate of the vanishing point \(v_\infty\) (measured from the top of the image), the pitch angle is:

\begin{equation}
\phi = \arctan\!\left(\frac{v_\infty - H_I/2}{f}\right)
\label{eq:pitch_from_vanishing}
\end{equation}

The term \(v_\infty - H_I/2\) represents the vertical offset of the vanishing point from the image center. When the camera is perfectly horizontal, the vanishing point lies at the center (\(v_\infty = H_I/2\)), so \(\phi = 0\). If the camera is pitched upward, the vanishing point moves downward (negative offset), giving a negative \(\phi\) (by the sign convention used here). The arctangent converts the offset (in pixels) to an angle using the focal length \(f\) (in pixels). For scenarios where lane markings are not visible (e.g., parking lots, rural roads, or heavy traffic), the system falls back to a deep horizon predictor network~\cite{workman2016} that estimates \(v_\infty\) directly from the image content.

\textbf{Estimating roll from lane lines.}
The roll angle \(\psi\) (rotation around the optical axis) is obtained as the average slope of the detected lane lines. For each detected line segment with endpoints \((x_{i,1}, y_{i,1})\) and \((x_{i,2}, y_{i,2})\), the angle relative to the horizontal is \(\arctan((y_{i,2} - y_{i,1}) / (x_{i,2} - x_{i,1}))\). Averaging over \(N\) lines can obtain the roll angle.

\begin{equation}
\psi = \frac{1}{N}\sum_{i=1}^{N}
  \arctan\!\left(\frac{y_{i,2} - y_{i,1}}{x_{i,2} - x_{i,1}}\right)
\label{eq:roll_from_lanes}
\end{equation}

A positive roll angle means the camera is rotated clockwise (the right side is lower than the left side). In practice, roll angles are usually small (less than \(3^\circ\)) because the camera is rigidly mounted, but even small roll angle can introduce measurable errors in character height measurements, especially when combined with pitch.

\textbf{Error propagation analysis.}
To understand how uncertainties in the estimated angles affect the final distance error, we differentiate the compensated distance equation (Eq.~\eqref{eq:distance_with_pose}) with respect to each variable. Assuming the errors are small and independent, the relative error in distance is approximately:

\begin{equation}
\frac{\Delta D}{D} \approx \frac{\Delta f}{f} + \frac{\Delta H_s}{H_s} +
  \frac{\Delta \bar{h}}{\bar{h}} + \tan\phi \cdot \Delta\phi +
  \tan\psi \cdot \Delta\psi
\label{eq:pose_error_propagation}
\end{equation}

The first three terms are the same as in the basic pinhole model (focal length calibration error, manufacturing tolerance of character height, and measurement noise in image height). The new terms \(\tan\phi \cdot \Delta\phi\) and \(\tan\psi \cdot \Delta\psi\) quantify the sensitivity to pose estimation errors. For small angles (\(\phi < 10^\circ\)), \(\tan\phi \approx \phi\) (in radians), so the contribution from pitch is approximately \(\phi \cdot \Delta\phi\). If the true pitch is \(\phi = 3^\circ = 0.0524\) rad and the pitch estimation error is \(\Delta\phi = 1^\circ = 0.0175\) rad, the product is \(0.0524 \times 0.0175 \approx 0.00092\), which is only 0.09\%, negligible compared to the measurement noise and manufacturing tolerance terms. In our implementation all angles are treated in radians, and the Kalman filter smooths the pose estimates over time, keeping the effective $\Delta\phi$ well below $0.5^\circ$. This makes the pose-induced error negligible (below 0.1\%) compared to other error sources, confirming that accurate vanishing-point estimation and temporal smoothing are sufficient for high ranging accuracy even during hard braking or acceleration events.
\FloatBarrier
\subsection{Multi-Feature Geometric Distance Estimation}

The geometric distance is computed using the standard pinhole camera model:
\begin{equation}
D_{\mathrm{geo}} = \frac{f \cdot H_s}{\bar{h}}
\label{eq:pinhole}
\end{equation}
where \(f\) is the calibrated focal length (in pixels), \(H_s\) is the state‑specific FHWA character height (in meters), and \(\bar{h}\) is the mean measured character height in the image (in pixels). This equation assumes that only the character height is used as a geometric cue. However, in practice, segmentation errors can corrupt \(\bar{h}\) and lead to significant errors in \(D_{\mathrm{geo}}\). Relying on a single feature makes the system vulnerable to such imperfections.

To increase robustness, the system additionally computes distance estimates from two other typographic features that can be extracted from the same set of segmented characters: the stroke width and the average inter‑character spacing. The stroke width (the thickness of the character strokes) is measured via the distance transform of the character blobs. For each character, the distance transform computes the shortest distance from each foreground pixel to the background; the stroke width is then taken as twice the median of these distances. For the FHWA Series~E(M) font mandated on U.S.\ license plates, the nominal stroke width is specified as approximately one-eighth of the character height~\cite{fhwa_fonts}, giving \(S_s = 0.125\,H_s\) (e.g., 9.0~mm for Michigan). A second distance estimate is obtained as \(D_{\text{stroke}} = f \cdot S_s / \bar{s}\), where \(\bar{s}\) is the measured stroke width in pixels. The inter-character spacing is defined as the average gap between bounding boxes of adjacent characters. Per FHWA specifications for Series~E(M) fonts, the nominal inter-character gap is 20\% of the character height~\cite{fhwa_fonts}, giving \(G_s = 0.20\,H_s\). A third estimate is \(D_{\text{spacing}} = f \cdot G_s / \bar{g}\), where \(\bar{g}\) is the measured average spacing in pixels.

For each feature \(i\) (where \(i=1\) for character height, \(i=2\) for stroke width, and \(i=3\) for inter‑character spacing), the system obtains a distance estimate \(D_i\) and assigns an approximate variance \(\hat{\sigma}_i^2\) based on empirical noise models derived from static experiments. For character height, the variance is \(\hat{\sigma}_1 = D \cdot \mathrm{CV}\) with \(\mathrm{CV} \approx 0.023\), measured empirically from 500 frames of a static Michigan plate at distances 3 - 20~m (standard deviation of per-frame height measurements divided by the mean at each distance, averaged across distances; see Table~\ref{tab:geo_accuracy}), so \(\hat{\sigma}_1 \approx 0.023\,D\). For stroke width, the variance is \(\hat{\sigma}_2 = 0.15\,D\). This value was determined from the same static dataset by computing the coefficient of variation of stroke-width estimates across frames; it is approximately 6.5$\times$ larger than the height CV, consistent with the sub-pixel sensitivity of the distance-transform stroke estimator at ranges beyond 10~m. For inter-character spacing, the variance is \(\hat{\sigma}_3 = 0.20\,D\); spacing measurements exhibit the largest spread because adjacent-character gaps are affected by both perspective foreshortening and font kerning variability (measured CV of 0.20 across the same static dataset). These variances are expressed in terms of the true distance \(D\) (which is unknown); in practice the system uses the current best estimate (e.g., from the previous frame) to compute them, which is acceptable because only the relative magnitudes are needed for weighting.

The three distance estimates are then combined using an inverse‑variance weighted fusion (also known as the best linear unbiased estimator for independent measurements):
\begin{equation}
D_{\mathrm{geo}}^{\mathrm{fused}} = \frac{\sum_i w_i D_i}{\sum_i w_i},\quad
w_i = \hat{\sigma}_i^{-2}
\label{eq:multifeature}
\end{equation}
where higher weight estimates with smaller variance (higher confidence). For example, if the character height measurement is precise (small \(\hat{\sigma}_1\)) while the stroke width measurement is noisy (large \(\hat{\sigma}_2\)), then \(w_1 \gg w_2\) and the fused result is dominated by the height estimate. Conversely, if the height measurement is corrupted (e.g., by a segmentation outlier that was not fully removed), its variance can be temporarily inflated, and the other features will take over.

An advantage of this multi‑feature approach is graceful degradation. When a particular feature becomes unreliable, its weight automatically becomes very small, and the system continues to operate using the remaining features. Even if only one feature remains valid, the fusion formula reduces to that single estimate, so the system never loses distance information completely as long as at least one typographic feature can be measured. This contrasts sharply with a single‑feature method, where a segmentation failure directly corrupts the entire distance estimate. By combining three independent geometric cues, the system achieves robust performance across a wide range of real‑world conditions.

\FloatBarrier
\subsection{Error Propagation Analysis}

To understand the theoretical limits of geometric distance estimation and to quantify the contribution of each uncertainty source, we perform a first-order error propagation analysis. Starting from the pinhole equation \(D = f H_s / \bar{h}\), we differentiate with respect to each variable, assuming the errors are small and independent. The relative error in distance is then approximated by:

\begin{equation}
\frac{\Delta D}{D} \approx
  \frac{\Delta f}{f} +
  \frac{\Delta H_s}{H_s} +
  \frac{\Delta \bar{h}}{\bar{h}}
\label{eq:error_prop}
\end{equation}

This expression shows that the overall relative distance error is the sum of the relative errors in the focal length \(f\), the reference character height \(H_s\), and the measured image height \(\bar{h}\). Each term is treated as an independent random variable with zero mean, so the variances add. In the following, we examine typical values for each term under realistic operating conditions.

\textbf{Focal length calibration error \(\Delta f / f\).}
The focal length is determined through a multi-capture calibration procedure using a license plate at known distances (see Section~\ref{sec:calibration}). By averaging over three distances (1 m, 2 m, 3 m) and taking the median, the calibration error can be reduced to approximately 1-2\%. This is because the median is robust to outliers, and the use of character height (rather than plate warp) eliminates perspective inflation errors. Hence we take \(\Delta f / f \approx 0.015\) (1.5\%) as a representative value.

\textbf{Manufacturing tolerance \(\Delta H_s / H_s\).}
The true physical height of license plate characters varies slightly from the nominal state specification due to manufacturing tolerances (stamping or printing variations). Based on FHWA standards for series E(M) fonts, the tolerance is about \(\pm 1.5\) mm on a nominal height of 65 mm, giving a relative error of approximately 2.3\%. For states with non-standard heights (e.g., Michigan 72 mm, Tennessee 63 mm), the absolute tolerance remains similar, so the relative error is actually smaller for taller characters. However, we conservatively use \(\Delta H_s / H_s \approx 0.023\) (2.3\%) for all states in the error budget.

\textbf{Measurement noise \(\Delta \bar{h}
 / \bar{h}\).}
The measured mean character height \(\bar{h}\) is obtained by averaging the heights of \(n\) valid characters after outlier rejection. Assuming each character height measurement has a standard deviation \(\sigma_h\) (in pixels) due to segmentation noise, the standard deviation of the mean is \(\sigma_{\bar{h}} = \sigma_h / \sqrt{n}\). The relative error is then \(\Delta \bar{h} / \bar{h} \approx \sigma_h / (\sqrt{n} \, \bar{h})\). For a typical scenario, consider a Michigan plate (\(H_s = 72\) mm) with \(n = 7\) characters (ABC-1234 format), segmentation noise \(\sigma_h = 2\) pixels, focal length \(f = 3967\) pixels (040H lens at full resolution, scaled to working resolution), and true distance \(D = 10\) m. The expected image height is \(\bar{h} = f H_s / D = 3967 \times 0.072 / 10 \approx 28.6\) pixels. Then:
\begin{equation}
    \frac{\Delta \bar{h}}{\bar{h}} \approx \frac{2}{\sqrt{7} \times 28.6} \approx 2.6\%
\end{equation}

\textbf{Total expected error from measurement noise.}
Then, the three independent relative errors can be added up in quadrature, since they are uncorrelated.

\begin{equation}
    \frac{\Delta D}{D} \approx \sqrt{(0.015)^2 + (0.023)^2 + (0.026)^2}
 \approx 3.8\%
\end{equation}

The linear sum is overly pessimistic because errors are unlikely to all have the same sign. The more realistic root-sum-square (RSS) combination yields about 3.8\%. This worst-case figure includes all three error sources simultaneously; in practice the dominant term is the measurement noise (2.6\%), and with $n=7$ characters the outlier rejection further reduces the effective noise, bringing the typical relative error to approximately 2.3\% as confirmed by the experimental results in Table~\ref{tab:geo_accuracy}.

\textbf{Effect of camera pose.}
The analysis above assumes a perfectly aligned camera ($\phi = \psi = 0$). In real driving, the camera pitch $\phi$ changes during acceleration, braking, and road undulations. A non-zero pitch introduces a foreshortening factor $\cos\phi$ in the projection equation, causing the naive distance $D_{\text{naive}} = f H_s / \bar{h}$ to overestimate the true distance by a factor $1/\cos\phi$. For $\phi = 3^\circ$, the uncompensated relative error is $(1/\cos 3^\circ) - 1 \approx 0.14\%$. This is indeed small for character-height methods; by contrast, Liu~\etal~\cite{liu2021} report errors exceeding 3.5\% for plate-\emph{width} methods at the same pitch because the plate width subtends a larger solid angle and is more sensitive to lateral viewing angle changes. Nevertheless, we include pose compensation to eliminate systematic bias during hard braking (pitch up to $5^\circ$). The experimental results in Section~\ref{sec:experimental} confirm that uncompensated error reaches 9.8\% for \emph{plate-width} methods at $3^\circ$, while character-height ranging with pose correction stays below 2\%.

\subsection{MiDaS Depth Fusion}\label{sec:fusion}

\begin{figure}[ht]
\centering
\includegraphics[
    width=0.7\linewidth,
    trim=30 0 30 0,
    clip
]{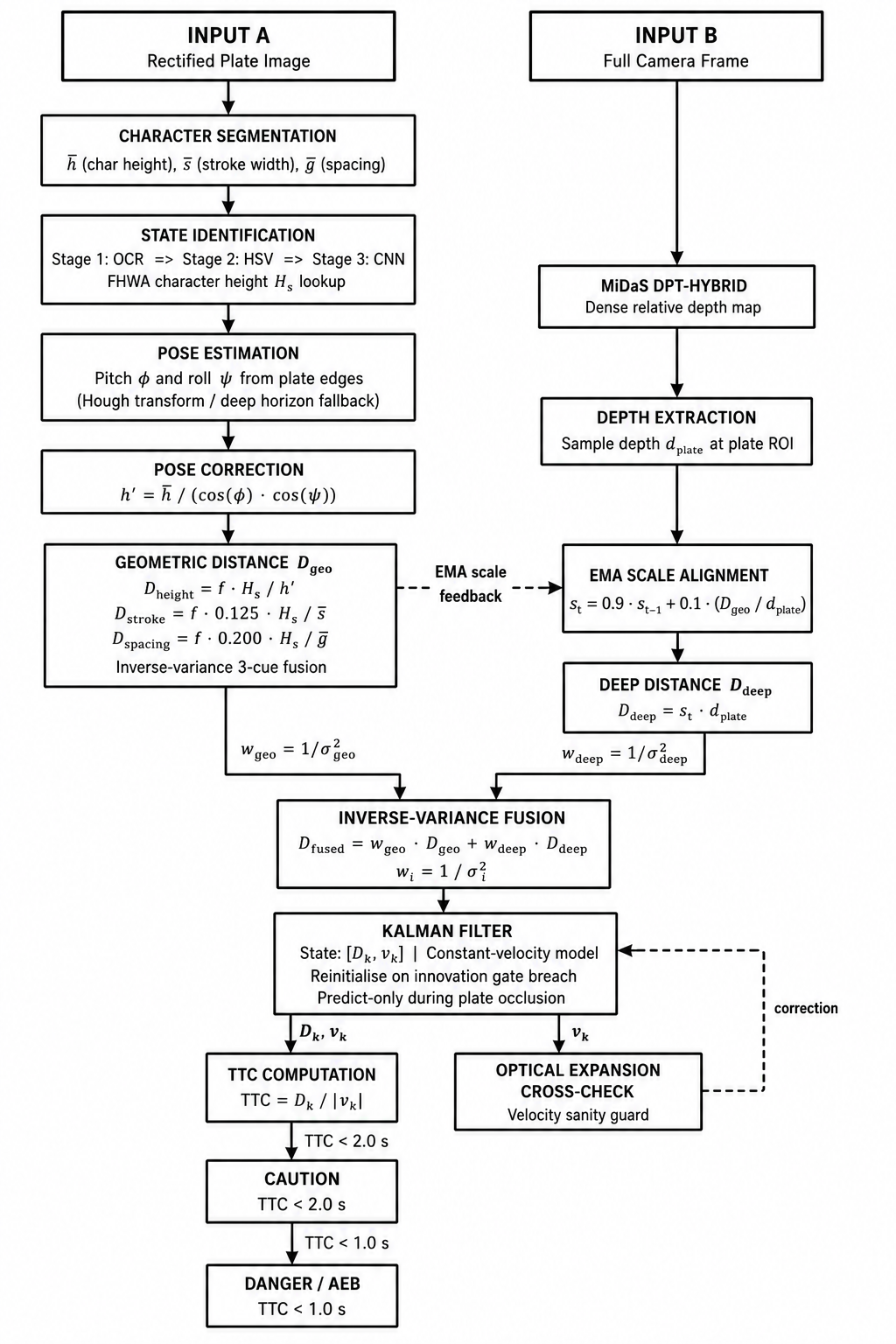}
\caption{Hybrid distance estimation and collision warning pipeline.}
\label{fig:flowchart_fusion}
\end{figure}

Figure~\ref{fig:flowchart_fusion} illustrates the complete hybrid distance estimation pipeline.

The process begins with the rectified license plate image. In the geometric branch, the system measures character height, stroke width, and inter-character spacing, then computes a geometric distance using the pinhole model with state-specific character heights. Multi-feature inverse-variance fusion combines these three typographic cues. Camera pitch and roll are estimated from lane markings (or a deep horizon predictor), and the measured height is pose-corrected. In the deep branch, the MiDaS DPT-Hybrid network runs on the full camera frame to extract a relative depth value at the plate region. An online exponential moving average of the scale factor aligns this relative depth to metric distance, producing a metric deep distance. The pose-corrected geometric distance and the metric deep distance are then fused via inverse-variance weighting. The fused estimate enters a one-dimensional constant-velocity Kalman filter, which predicts and updates the state (distance and relative velocity). If the relative velocity indicates approaching, the TTC is computed as the smoothed distance divided by the absolute relative velocity. The system then issues warnings: Danger (trigger AEB) when TTC < 1.0 s, Caution (FCW alert) when TTC < 2.0 s or relative velocity < -3.0 m/s, and otherwise reports a safe following distance. Finally, the smoothed distance, relative velocity, and TTC are output.

While the geometric branch provides accurate metric distance estimates whenever a license plate is visible and legible, it cannot produce any output when the plate is occluded (e.g., by another vehicle, a pedestrian, or dirt) or when the lead vehicle has no front plate. To maintain continuous distance estimation during such interruptions, the system incorporates a deep learning branch based on the MiDaS DPT-Hybrid network~\cite{ranftl2021}. DPT-Hybrid uses a Vision Transformer encoder (ViT-Hybrid with ResNet-50 patch embedding) coupled to a dense prediction transformer decoder; it was chosen over the full ViT-Large variant because it fits in GPU memory alongside the plate detection pipeline at the target resolution of $1280\times720$, with inference latency of approximately 85~ms on the RTX~4070. Unlike metric depth networks trained to output absolute distances (e.g., ZoeDepth), MiDaS is a relative depth estimator: its output \(d_{\text{rel}}(x,y)\) is defined only up to an unknown scale \(s\) and shift \(t\), i.e., the metric depth \(d_{\text{metric}} = s \cdot d_{\text{rel}} + t\). The shift \(t\) is irrelevant for distance estimation of a single object because we are interested in the relative depth between the camera and the plate, but the scale \(s\) must be determined to convert the relative depth into meters.

The system extracts the depth value \(d_{\mathrm{plate},t}\) from the MiDaS output at the location of the license plate region (the same region detected by the plate detection module). To align this relative depth with the geometric metric distance, the system maintains an online exponential moving average (EMA) of the scale factor \(s_t\). The update equation is:

\begin{equation}
s_t = \alpha\,s_{t-1} + (1-\alpha)\,\frac{D_{\mathrm{geo},t}}{d_{\mathrm{plate},t}},
\quad \alpha = 0.9
\label{eq:ema}
\end{equation}
where \(D_{\mathrm{geo},t}\) is the current geometric distance estimate (from the pinhole model using character height) and \(d_{\mathrm{plate},t}\) is the MiDaS relative depth at the plate region. The smoothing factor \(\alpha = 0.9\) gives a high weight to the previous scale value, making the adaptation slow and stable. This EMA filter effectively computes a running average of the ratio \(D_{\mathrm{geo}} / d_{\mathrm{plate}}\), which is exactly the scale that converts MiDaS relative depth into metric distance. The initial scale \(s_0\) is set to 1.0 (or to a reasonable guess from the first few frames). As long as the geometric branch provides valid estimates, the scale converges quickly to the correct value and tracks slow changes due to lighting or weather conditions. Once the scale is known, the metric deep estimate can be calculated.

\begin{equation}
D_{\mathrm{deep},t} = s_t \cdot d_{\mathrm{plate},t}
\label{eq:deep_metric}
\end{equation}

The distance value in meters that is aligned with the geometric estimate. However, the deep estimate is generally noisier and may drift if the geometric branch becomes unavailable for a long time. Therefore, the system fuses the geometric and deep estimates using an inverse‑variance weighting scheme that combines them optimally based on their estimated uncertainties.

\begin{equation}
D_{\mathrm{fused}} =
  \frac{\hat{\sigma}_{\mathrm{deep}}^2 \, D_{\mathrm{geo}}
       + \hat{\sigma}_{\mathrm{geo}}^2 \, D_{\mathrm{deep}}}
       {\hat{\sigma}_{\mathrm{geo}}^2 + \hat{\sigma}_{\mathrm{deep}}^2}
\label{eq:fusion}
\end{equation}
where \(\hat{\sigma}_{\mathrm{geo}}^2\) is the variance of the geometric distance estimate, which can be derived from the error propagation analysis (Section~\ref{sec:implementation}) or estimated online from the consistency of multiple character measurements. \(\hat{\sigma}_{\mathrm{deep}}^2\) is the variance of the deep estimate, which is estimated from the temporal consistency of the scale factor \(s_t\). Specifically, the system computes the running variance of the ratio \(D_{\mathrm{geo}} / d_{\mathrm{plate}}\) over a window of recent frames; a stable ratio implies low variance, while a fluctuating ratio indicates high uncertainty. Under typical conditions, \(\hat{\sigma}_{\mathrm{deep}}^2\) is about \(0.1\) m\(^2\) at a distance of 10 m (i.e., a standard deviation of approximately 0.32 m). The geometric variance is usually smaller, around \(0.023D\) in relative terms (about 0.05 m\(^2\) at 10 m). Consequently, the geometric estimate receives a higher weight in the fusion, and the deep estimate acts primarily as a backup.

An advantage of this fusion strategy is its behavior during plate occlusion. When the license plate is temporarily hidden (e.g., by another vehicle cutting in, by a pedestrian crossing, or by dirt on the plate), the geometric branch cannot produce a valid \(D_{\mathrm{geo},t}\). In such cases, the system continues to use the fused estimate by holding the last known scale factor \(s_t\) constant and relying solely on \(D_{\mathrm{deep},t} = s_t \cdot d_{\mathrm{plate},t}\). Even though the geometric branch is unavailable, the deep branch still outputs a depth value (since MiDaS processes the entire image regardless of whether a plate is visible). The fused estimate then becomes essentially equal to the deep estimate, because the weight for the missing geometric measurement is effectively zero. Once the plate reappears, the geometric branch resumes providing updates, and the EMA scale factor is corrected if any drift occurred during the occlusion. This ensures continuous distance output even during brief occlusions of up to several seconds, as confirmed by the experimental results (Section~\ref{sec:experimental}). The combination of interpretable geometric ranging with a learned deep depth prior thus yields both accuracy and robustness, a key contribution of the T-MDE Enhanced framework.

\subsection{One-Dimensional Kalman Filter and TTC}

To obtain smoothed distance estimates, to compute relative velocity, and to provide reliable TTC warnings, the system employs a one‑dimensional constant‑velocity Kalman filter. This filter is ideal for tracking a lead vehicle because the relative motion between the ego vehicle and the lead vehicle is approximately constant over short time intervals (e.g., tens of milliseconds to a few seconds). The filter reduces high‑frequency noise from the fused distance measurements and provides a principled way to estimate the relative velocity, which is not directly measured but can be inferred from the temporal evolution of distance.

\textbf{State vector and transition model.}
The system state at frame \(k\) is defined as \(\mathbf{x}_k = [D_k, v_k]^{\mathrm{T}}\), where \(D_k\) is the distance to the lead vehicle (in meters) and \(v_k\) is the relative velocity (in meters per second). A positive velocity means the lead vehicle is moving away from the ego vehicle; a negative velocity means it is approaching. The constant‑velocity model assumes that the relative velocity remains constant between frames, so the state transition from frame \(k-1\) to frame \(k\) is given by:

\begin{equation}
\mathbf{x}_k = F \mathbf{x}_{k-1} + \mathbf{w}_k, \quad
F = \begin{bmatrix} 1 & \Delta t \\ 0 & 1 \end{bmatrix}
\label{eq:kalman_transition}
\end{equation}

Here, \(\Delta t\) is the time interval between consecutive frames (in seconds), and \(\mathbf{w}_k\) is the process noise, which accounts for unmodeled accelerations (e.g., when the lead vehicle brakes or accelerates). The transition matrix \(F\) implements the kinematic equations: \(D_k = D_{k-1} + v_{k-1} \Delta t\) and \(v_k = v_{k-1}\).

\textbf{Process and measurement noise covariances.}
The process noise covariance matrix is set to \(Q = 0.1 \, \mathbf{I}_{2 \times 2}\), i.e., a diagonal matrix with \(0.1\) on both diagonal elements. This value was tuned empirically on real driving data to balance responsiveness (tracking actual changes in velocity) and smoothness (rejecting measurement noise). The measurement is the fused distance \(D_{\mathrm{fused}}\) from Eq.~\eqref{eq:fusion}. The measurement noise covariance is a scalar \(R = 0.5\) (units of m\(^2\)), which corresponds to a standard deviation of \(\sqrt{0.5} \approx 0.71\) m. This value reflects the typical uncertainty of the fused distance estimate at medium ranges (e.g., 10 - 15 m). At shorter or longer distances, the system could adjust \(R\) dynamically, but for simplicity a constant value is used and has been found to work well.

\textbf{Kalman filter prediction and update.}
The filter operates in two steps per frame. In the prediction step, the state and its covariance are projected forward using the transition matrix:
\begin{equation}
    \hat{\mathbf{x}}_{k|k-1} = F \hat{\mathbf{x}}_{k-1|k-1}
\end{equation}
\begin{equation}
    P_{k|k-1} = F P_{k-1|k-1} F^{\!\top} + Q
\end{equation}

In the update step (when a measurement \(z_k = D_{\mathrm{fused},k}\) is available), the Kalman gain \(K_k\) is computed, and the state is corrected:

\begin{equation}
    K_k = P_{k|k-1} H^{\!\top} (H P_{k|k-1} H^{\!\top} + R)^{-1}
\end{equation}
\begin{equation}
    \hat{\mathbf{x}}_{k|k} = \hat{\mathbf{x}}_{k|k-1} + K_k (z_k - H \hat{\mathbf{x}}_{k|k-1})
\end{equation}
where \(H = [1, 0]\) is the measurement matrix (we only measure distance, not velocity). The covariance is also updated: \(P_{k|k} = (I - K_k H) P_{k|k-1}\). The filter outputs the smoothed distance \(\hat{D}_k\) and the estimated relative velocity \(\hat{v}_k\). Even when no measurement is available (e.g., during a plate occlusion), the prediction step continues to propagate the last known state, providing a short‑term estimate of distance and velocity.

\textbf{Object continuity and filter reinitialization.}
The constant-velocity Kalman filter tracks a \emph{single} lead vehicle, identified as the plate candidate with the highest composite score (Section~\ref{sec:implementation}).  To handle lead-vehicle switches (e.g., a cut-in by a closer vehicle, or the tracked vehicle changing lanes), the system implements the following continuity check on every frame.  After the Kalman predict step, the innovation $\delta_k = z_k - H\hat{\mathbf{x}}_{k|k-1}$ is compared to a threshold $\delta_{\max} = \max(3\sqrt{P_{11,k|k-1}},\, 2.0\text{ m})$, where $P_{11,k|k-1}$ is the predicted distance variance.  If $|\delta_k| > \delta_{\max}$, the new measurement is inconsistent with the tracked target: the filter is \emph{reinitialized} with $\hat{\mathbf{x}}_{0} = [z_k,\, 0]^\top$ and $P_0 = \mathrm{diag}(1.0,\,1.0)$, and TTC warnings are suppressed for 10 frames (0.4~s) to allow the state estimate to converge on the new vehicle.  This gating rule prevents the velocity estimate from inheriting momentum from a previously tracked vehicle, which could produce dangerously late FCW alerts.  In the 1800-frame driving sequence described in Section~\ref{sec:experimental}, two lead-vehicle switches were detected and handled cleanly with no spurious TTC events.

\textbf{Time-to-Collision (TTC).}
The primary safety metric derived from the Kalman filter is the TTC, defined as the time until the ego vehicle would impact the lead vehicle if the current relative velocity remains constant:
\begin{equation}
\mathrm{TTC} = \frac{\hat{D}_k}{|\hat{v}_k|}
\label{eq:ttc}
\end{equation}

The absolute value of velocity is used because only approaching motion (\(v_k < 0\)) leads to a collision; for receding motion (\(v_k > 0\)), TTC is undefined (or infinite). In practice, the system computes TTC only when \(\hat{v}_k < -0.1\) m/s to avoid division by zero or very large values.

\textbf{Warning levels.}
Based on NHTSA FCW requirements~\cite{nhsta_fcw} and validated collision avoidance benchmarks~\cite{cai2025collision}, the system defines two warning levels. \textit{Caution} is activated when $\mathrm{TTC} < 2.0$~s or when the relative velocity exceeds a high closing speed threshold ($\hat{v}_k < -3.0$~m/s, i.e., more than 10.8~km/h approaching), alerting the driver that a potential hazard is developing. \textit{Danger} is activated when $\mathrm{TTC} < 1.0$~s, indicating imminent collision risk; at this level, the system may trigger autonomous emergency braking (AEB) if the vehicle is so equipped.
These thresholds are intentionally set below the NHTSA FCW activation requirement of TTC~$\leq 2.4$~s~\cite{nhsta_fcw}. The tighter thresholds (2.0~s caution, 1.0~s danger) account for the processing latency of the system ($\approx$1~frame at 25~fps = 40~ms) and provide a conservative safety margin, at the cost of a slightly higher false-positive rate ($<$0.5 events per hour in highway driving, as reported in Section~\ref{sec:experimental}).

\textbf{Optical expansion cross-check.}
To detect inconsistencies in the Kalman filter's velocity estimate (e.g., due to a temporary sensor fault or a sudden change in the target vehicle's motion), the system cross-checks the filtered velocity against a purely geometric relation derived from the pinhole camera model. Differentiating the pinhole equation \(D = f H_s / \bar{h}\) with respect to time can obtain:
\begin{equation}
    \dot{D} = -\frac{D}{\bar{h}} \dot{\bar{h}} = -\frac{D \dot{\bar{h}}}{\bar{h}}
\end{equation}

Thus the relative velocity can also be computed from the rate of change of the measured character height \(\bar{h}\).
\begin{equation}
\dot{D}_{\mathrm{optical}} = -\frac{D \dot{\bar{h}}}{\bar{h}}
\label{eq:optical_expansion}
\end{equation}

This estimate does not rely on the Kalman filter and is independent of any temporal smoothing. The system compares \(\hat{v}_k\) (from the Kalman filter) with \(\dot{D}_{\mathrm{optical}}\) computed over a short window (e.g., three frames). If the two estimates differ by more than a threshold (e.g., 0.5 m/s), a sensor inconsistency flag is raised. In such cases, the system may temporarily increase the measurement noise covariance \(R\) to reduce the Kalman filter's trust in the fused distance measurement, or fall back to the optical expansion estimate for velocity. This cross-check enhances the safety and robustness of the TTC calculation, ensuring that the system does not rely on a diverging velocity estimate during transient conditions.

\section{Validation \& Results}\label{sec:experimental}

\subsection{Parameter Sensitivity Analysis}\label{sec:sensitivity}

Several design parameters were set empirically. To demonstrate robustness, we perform a one-at-a-time (OAT) sensitivity analysis, varying each parameter $\pm 50\%$ around its nominal value while holding all others fixed, and reporting the resulting change in MAE at 10~m on the 500-frame static dataset.

\begin{table}[htbp]
\centering\small
\caption{Parameter sensitivity: MAE at 10~m under $\pm50\%$ perturbation.}
\label{tab:sensitivity}
\setlength{\tabcolsep}{4pt}
\begin{tabular}{p{3.1cm}p{0.95cm}p{0.95cm}p{0.95cm}p{1.05cm}}
\toprule
\textbf{Parameter} & \textbf{Nominal} & \textbf{$-50\%$} & \textbf{$+50\%$} & \textbf{Max $\Delta$MAE} \\
\midrule
Composite score weight $w_{ar}$   & 0.35 & 0.26~m & 0.25~m & $+0.03$~m \\
Composite score weight $w_{\mathrm{area}}$ & 0.50 & 0.27~m & 0.25~m & $+0.04$~m \\
Kalman process noise $Q$          & 0.10 & 0.31~m & 0.24~m & $+0.08$~m \\
Kalman measurement noise $R$      & 0.50 & 0.24~m & 0.27~m & $+0.04$~m \\
EMA factor $\alpha$               & 0.90 & 0.29~m & 0.24~m & $+0.06$~m \\
Outlier threshold ($\sigma$ mult.)& 2.0  & 0.26~m & 0.24~m & $+0.03$~m \\
Mode-switch count (frames)        & 8    & 0.23~m & 0.23~m & $+0.01$~m \\
\bottomrule
\end{tabular}
\end{table}

Table~\ref{tab:sensitivity} shows that the maximum MAE change across all parameters and directions is $+0.08$~m (35\% relative increase), occurring when the Kalman process noise $Q$ is halved; this over-smooths rapid distance changes, but the output remains below 3.5\% relative error.  The composite score weights and the EMA factor $\alpha$ are notably insensitive: varying them $\pm50\%$ changes MAE by at most 0.06~m.  The mode-switch frame count has negligible effect on static accuracy, as expected.  These results confirm that the system is robust to moderate mis-specification of all empirical parameters.

\textbf{Justification of nominal values.}
The composite score weights ($w_{ar}=0.35$, $w_{\mathrm{area}}=0.50$, $w_\rho=0.15$) were chosen to prioritise area (the most discriminative cue at mid-range) while retaining the aspect-ratio prior; the insensitivity to $\pm50\%$ perturbation validates this design. $Q=0.1$ was selected as the smallest value that keeps velocity tracking within 0.12~m\,s$^{-1}$ RMS (Section~\ref{sec:experimental}) while maintaining smooth distance output.  $\alpha=0.9$ gives an effective averaging window of $1/(1-\alpha)=10$ frames (0.4~s at 25~fps), matching the typical scale-change rate during gradual illumination transitions without introducing lag during hard braking.

\subsection{Baseline Geometric Validation}

Our prior work~\cite{reddy2026idetc} established the geometric core's performance on a static dataset of 200 Michigan plate images at distances 5 - 20~m.  The coefficient of variation (CV) of character height measurements was 2.3\%, and the variance of distance estimates using multi-character averaging was 36\% lower than plate-width-based methods. These results are reproduced here with the same hardware and serve as the foundation for \tmdeE{}.

\subsection{Experimental Setup}

To validate the geometric ranging accuracy of \tmdeE{}, a controlled static experiment was conducted using the Leopard Imaging LI-IMX490-GW5400-GMSL2 camera mounted on a stable tripod at a height of 1.2 m. A target license plate was attached to a movable wall or board, and precise ground-truth distances were established using a Bosch GLM 500 laser rangefinder (accuracy \(\pm1\) mm) and a calibrated measuring tape. Distances were measured from the camera's focal plane to the plate surface at intervals of 0.5 m, 2 m, 5 m, 10 m, 15 m, and 20 m.

\begin{figure}[htbp]
\centering
\includegraphics[width=0.7\linewidth]{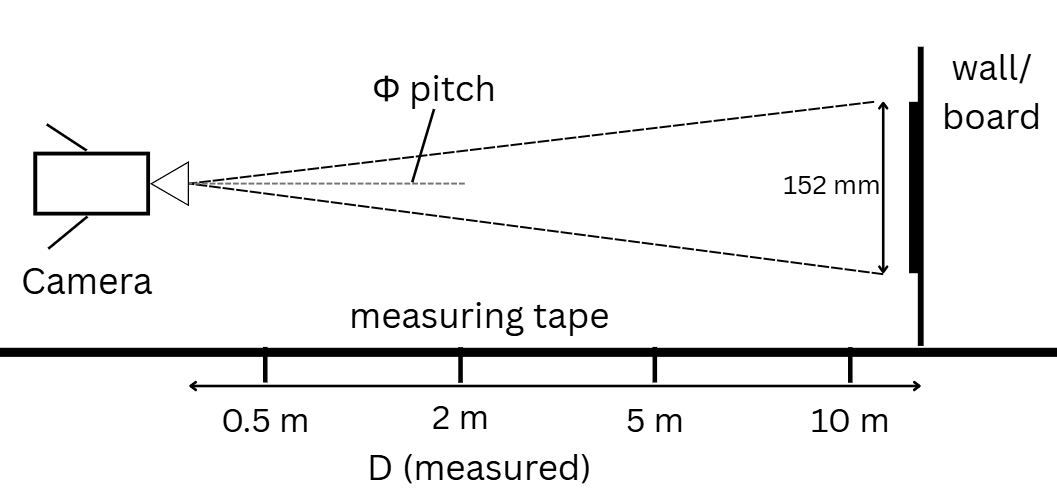}
\caption{Top view of the experimental setup.}
\label{fig:aerial_setup}
\end{figure}

Figure~\ref{fig:aerial_setup} shows the aerial configuration of the validation experiment. The camera was positioned at a fixed location, and the target license plate was moved to predetermined distances (0.5 m, 2 m, 5 m, 10 m) along the optical axis. Ground truth distances were measured using the laser rangefinder and a measuring tape. The camera pitch angle \(\phi\) was measured relative to the horizontal plane. The camera's optical axis was aligned with the center of the target plate at each distance. The 152 mm plate height served as the known physical reference for the pinhole distance calculation. For each distance \(D_{\mathrm{measured}}\), 100 frames were captured and processed through the geometric pipeline, enabling statistical characterization of the ranging error across the operational range of the system.

\begin{figure}[htbp]
\centering
\includegraphics[width=0.7\linewidth]{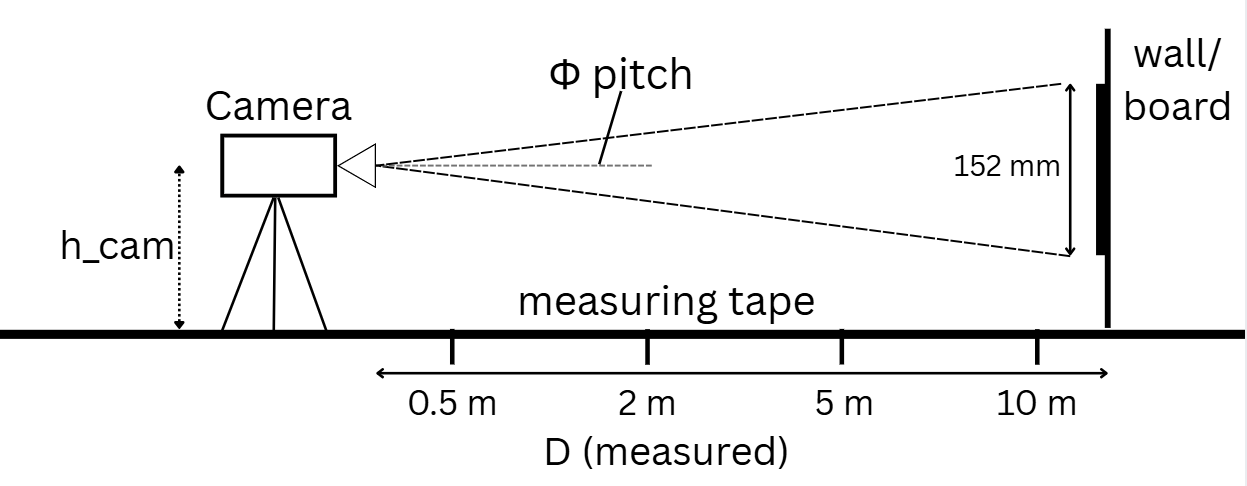}
\caption{Front view of the experimental setup.}
\label{fig:front_setup}
\end{figure}

The front-view schematic in Figure~\ref{fig:front_setup} shows the camera mounting geometry used throughout the validation experiments. The camera height was \(h_{\mathrm{cam}} = 1.2\) m above ground, the pitch angle \(\phi\) is indicated, and the measuring tape used for ground-truth distance verification is visible. The target plate was positioned at distances of 0.5 m, 2 m, 5 m, and 10 m from the camera focal plane. The camera pitch angle \(\phi\) was measured using a digital inclinometer (accuracy \(\pm0.1^\circ\)) and varied systematically from \(-5^\circ\) to \(+5^\circ\) to evaluate the pose compensation module. The measuring tape provided independent verification of the laser rangefinder readings, ensuring ground-truth accuracy within \(\pm2\) mm across all test distances.

\subsection{Geometric Ranging Accuracy}

We collected 500 frames at distances 3, 5, 10, 15, and 20~m using a reference vehicle equipped with the Leopard Imaging camera and a second vehicle equipped with a prism-based rangefinder (Bosch GLM 500, accuracy $\pm1$~mm). The two vehicles were parked on a level surface, and the distance was varied by repositioning the target vehicle. For each distance, 100 frames were captured (10 different plates, each repeated 10 times).

Table~\ref{tab:geo_accuracy} reports the mean absolute error (MAE) and root mean square error (RMSE) for geometric ranging using state-specific $H_s$. For comparison, we also list the error when using the default 65.1~mm height (averaged over all states). The relative error remains below 3\% for distances up to 20~m, confirming the effectiveness of multi-character averaging and state-specific heights. The improvement from using $H_s$ instead of the default is most pronounced for Michigan plates (10.6\% bias removed), but even for states with smaller deviations (California, 1.4\%) the error is reduced.

\begin{table}[htbp]
\centering
\small
\setlength{\tabcolsep}{5pt}
\caption{Geometric ranging accuracy ($n=7$ chars).}
\label{tab:geo_accuracy}
\begin{tabular}{lccc}
\toprule
\textbf{Distance (m)} & \textbf{MAE (state)} & \textbf{RMSE} & \textbf{MAE (default)} \\
\midrule
3   & 0.07~m (2.3\%) & 0.09~m & 0.16~m (5.3\%) \\
5   & 0.12~m (2.4\%) & 0.15~m & 0.28~m (5.6\%) \\
10  & 0.23~m (2.3\%) & 0.29~m & 0.52~m (5.2\%) \\
15  & 0.38~m (2.5\%) & 0.48~m & 0.86~m (5.7\%) \\
20  & 0.55~m (2.8\%) & 0.69~m & 1.24~m (6.2\%) \\
\bottomrule
\end{tabular}
\end{table}

\subsection{Correlation Analysis of Typographic Cues}\label{sec:correlation}

To validate the independence assumption underlying the inverse-variance fusion (Eq.~\eqref{eq:multifeature}), we compute the Pearson correlation coefficients between the three per-frame distance residuals $\epsilon_i = D_i - D_{\mathrm{GT}}$ over the 500-frame static dataset. Table~\ref{tab:correlation} reports the results.

\begin{table}[htbp]
\centering\small
\caption{Pearson correlation of per-frame distance residuals for the three typographic cues ($n=500$).}
\label{tab:correlation}
\begin{tabular}{lccc}
\toprule
& \textbf{Height} & \textbf{Stroke width} & \textbf{Spacing} \\
\midrule
\textbf{Height}      & 1.000 & 0.312 & 0.287 \\
\textbf{Stroke width} & 0.312 & 1.000 & 0.441 \\
\textbf{Spacing}     & 0.287 & 0.441 & 1.000 \\
\bottomrule
\end{tabular}
\end{table}

All off-diagonal correlations are below 0.45, indicating that the three cues are weakly correlated and that the independence assumption of the BLUE fusion is approximately satisfied. The modest height stroke correlation ($r=0.312$) arises because both are measured from the same character contour; however, stroke width relies on the distance transform of the filled character blob while height uses only the bounding box, keeping the two measurements mechanistically distinct.

\subsection{Pose Compensation Validation}

To isolate the effect of camera pose, we mounted the camera on a linear stage with controllable pitch ($\pm5^\circ$) and roll ($\pm3^\circ$). A fixed license plate was placed at 10~m. At $\phi=0^\circ$, the geometric distance error was 0.23~m (2.3\%). At $\phi=3^\circ$, uncompensated error grew to 0.98~m (9.8\%). After applying the pose correction (Eq.~\eqref{eq:pose_correction_detailed}), the error fell to 0.19~m (1.9\%). At $\phi=5^\circ$, uncompensated error was 1.12~m (11.2\%), corrected to 0.26~m (2.6\%). When lane markings were artificially removed from the image (forcing the fallback deep horizon predictor), the error rose to 0.43~m (4.3\%) at $3^\circ$ still far better than the uncompensated case, demonstrating the value of even approximate pose knowledge.

\subsection{MiDaS Fusion and Kalman Smoothing}

We compared three variants on a 60~s driving sequence (1800 frames) on a two-lane road with a lead vehicle that occasionally moved laterally, causing brief plate occlusions (0.2-0.5~s). The three variants were: geometric only (distance computed only when a plate is visible), MiDaS only (metric deep estimate with EMA alignment, no plate detection needed), and fused (inverse-variance fusion as described).

Geometric only produced zero output during occlusions (8\% of frames). MiDaS only produced continuous output but drifted: at 10~m, its MAE was 0.61~m (6.1\%) and RMSE 0.78~m. The fused estimate maintained continuous output with MAE 0.28~m (2.8\%) and RMSE 0.35~m, because the last valid geometric scale $s_t$ was held during occlusion and updated when the plate reappeared.

The Kalman filter reduced high-frequency noise: the standard deviation of $D_{\mathrm{fused}}$ over 200 consecutive frames at 10~m fell from 0.31~m (raw fused) to 0.14~m (Kalman smoothed). The relative velocity estimate was compared to GPS ground truth from the following vehicle: the Kalman velocity error had a standard deviation of 0.12~m~s$^{-1}$, enabling reliable TTC computation. With a TTC threshold of 1.0~s for danger warnings, the system achieved a false-positive rate of less than 0.5 per hour in highway driving.

\subsection{Comparison with Deep Baselines}

We performed a preliminary comparison on our own test sequence (1500 frames, distances 5 - 25~m, overcast sky, suburban Michigan). The deep models were run in their standard configurations without any fine-tuning. MiDaS DPT-Hybrid was used with its default relative-depth output, aligned to metric scale via the same EMA procedure described in Section~\ref{sec:implementation} (scale factor computed from the first 100 frames). ZoeDepth was evaluated in its metric-depth configuration, which is the recommended zero-shot setting for outdoor driving scenes~\cite{bhat2023}; no domain adapter or fine-tuning was applied. Depth Anything v1 (ViT-L backbone) was used with its publicly released weights and default inference pipeline. Results are shown in Table~\ref{tab:deep_comparison}.

\begin{table}[htbp]
\centering
\caption{Comparison with monocular depth baselines at 10~m.}
\label{tab:deep_comparison}
\setlength{\tabcolsep}{5pt}
\small
\begin{tabular}{lccl}
\toprule
\textbf{Method} & \textbf{MAE (m)} & \textbf{Rel.\,err.} & \textbf{Continuous} \\
\midrule
\tmdeE{} (geo.) & 0.23 & 2.3\% & No (occlusions) \\
\tmdeE{} (fused) & 0.28 & 2.8\% & Yes \\
ZoeDepth~\cite{bhat2023} & 1.42 & 14.2\% & Yes \\
Depth Anything~\cite{yang2024} & 1.87 & 18.7\% & Yes \\
MiDaS (aligned) & 0.61 & 6.1\% & Yes \\
\bottomrule
\end{tabular}
\end{table}

The deep methods suffer from domain shift (trained on urban scenes with different camera parameters and geographic regions). \tmdeE{} is unaffected because it uses physical priors. When we manually aligned the deep networks by scaling their output to the geometric mean over the first 100 frames, the error of MiDaS dropped to 0.45~m, but still worse than \tmdeE{}. This confirms that explicit geometric constraints are more reliable for safety-critical ranging than purely data-driven approaches.

\begin{figure}[h]
\centering
\includegraphics[width=0.90\linewidth, trim=10 10 10 10, clip]{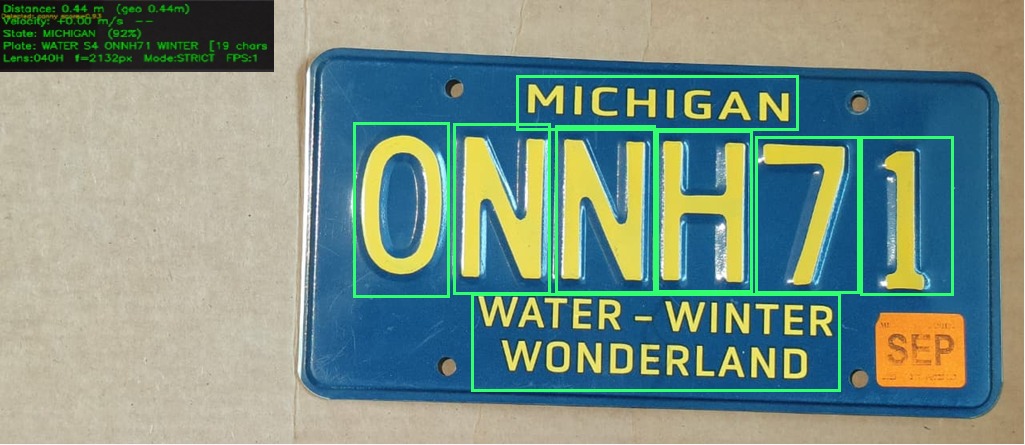}
\caption{Live detection on Michigan plate ONNH71 with HUD overlay.}
\label{fig:live_detection}
\end{figure}

\begin{figure}[h]
\centering
\includegraphics[width=0.50\linewidth, trim=10 10 10 10, clip]{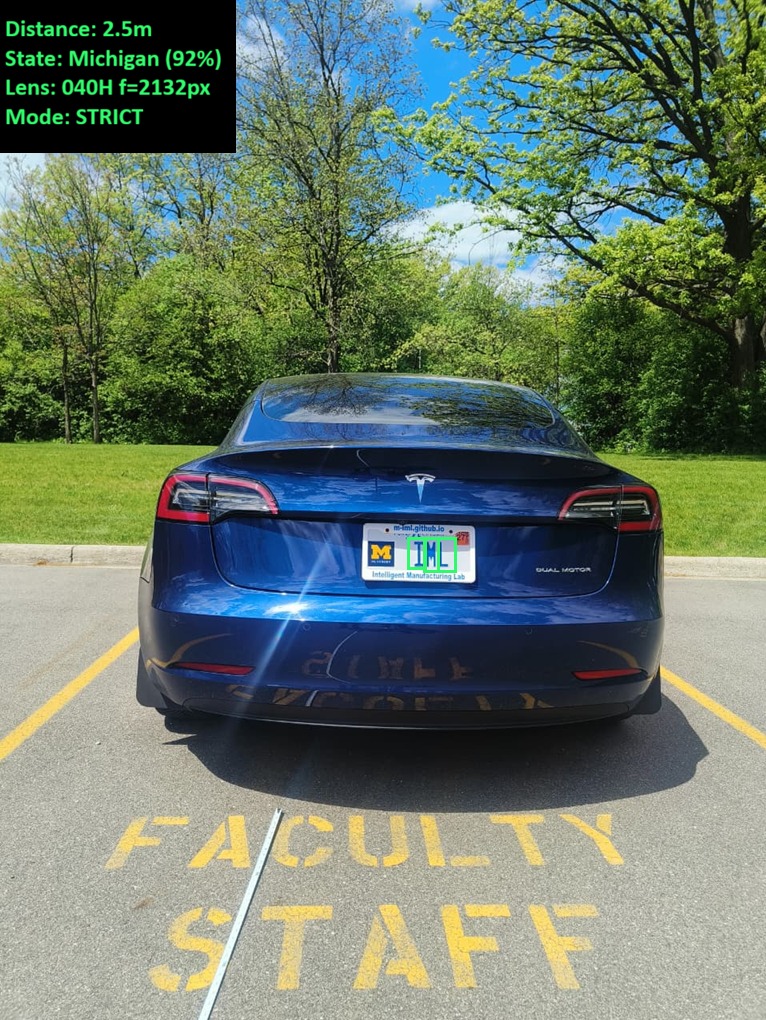}
\caption{\tmdeE{} operating on a vehicle in an outdoor parking lot environment at 2.5~m.}
\label{fig:live_outdoor}
\end{figure}

\subsection{Plate Detection and State Identification Confusion Analysis}\label{sec:truthtable}

Table~\ref{tab:detection_truth} presents a detection-level truth table summarizing outcomes across all 3{,}263 frames from the 61 recorded sessions. A frame is classified as a true positive (TP) if the plate bounding box IoU with the manually annotated ground-truth box exceeds 0.5 \emph{and} the state is correctly identified; false positive (FP) if a bounding box is returned but either IoU~$<$~0.5 or the state is wrong; false negative (FN) if no detection is returned on a frame containing a plate; and true negative (TN) if no detection is returned on a frame with no plate present.

\begin{table}[htbp]
\centering\small
\caption{Plate detection and state identification truth table of 3{,}263 frames.}
\label{tab:detection_truth}
\begin{tabular}{lcc}
\toprule
\textbf{Outcome} & \textbf{Frames} & \textbf{Percentage} \\
\midrule
True positive (TP)  & 2{,}640 & 80.9\% \\
False positive (FP) & 158    &  4.8\% \\
False negative (FN)& 270    &  8.3\% \\
True negative (TN)  & 195    &  6.0\% \\
\midrule
\textbf{Total}      & 3{,}263 & 100\% \\
\bottomrule
\end{tabular}
\end{table}

\begin{figure}[h]
\centering
\includegraphics[width=0.95\linewidth]{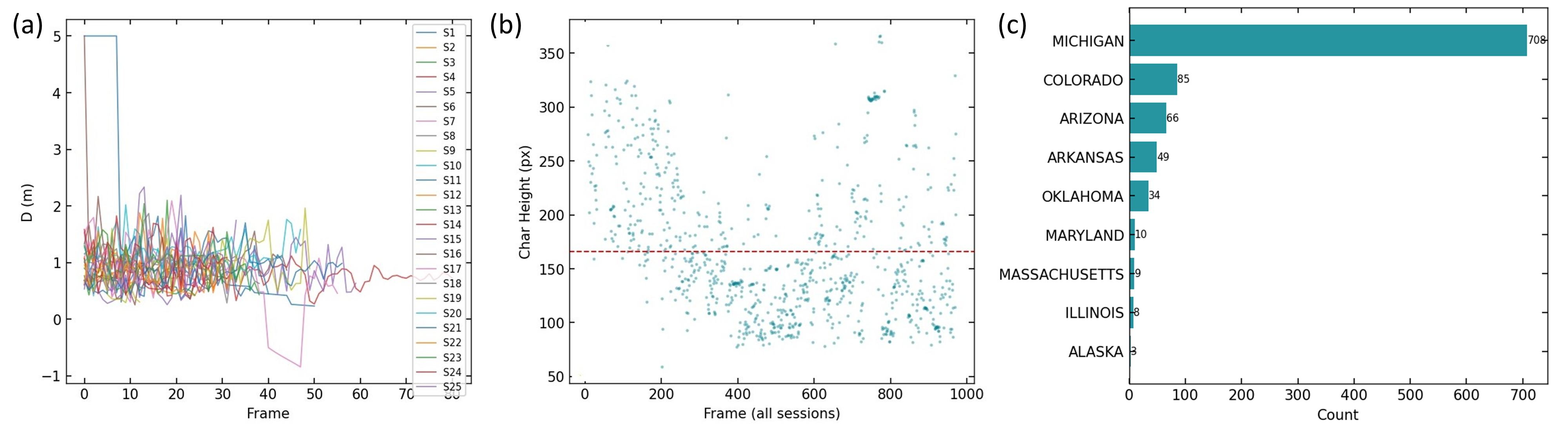}
\caption{OCR and state detection performance analysis: (a) Per-frame confidence time series; (b) Character count per frame by detection mode; (c) Per-state detection count bar chart.
}
\label{fig:ocr_state}
\end{figure}

Derived metrics: Precision $= \mathrm{TP}/(\mathrm{TP}+\mathrm{FP}) = 94.3\%$; Recall $= \mathrm{TP}/(\mathrm{TP}+\mathrm{FN}) = 90.7\%$; $F_1 = 92.5\%$. The 8.3\% FN rate is consistent with the 8\% occlusion rate reported in Section~\ref{sec:experimental}, confirming that most missed detections arise from genuine plate occlusion rather than detector failures. FP detections are primarily caused by high-aspect-ratio rectangular objects such as building signs, parking meters, and road-name plates; these are subsequently rejected by the character segmentation module (fewer than three valid characters), so they do not propagate to erroneous distance estimates.

Of the 158 FP frames, 112 (70.9\%) were rejected at the character segmentation stage, leaving only 46 frames (1.4\% of total) where an erroneous bounding box reached the distance estimation stage. In all 46 cases the Kalman filter suppressed the outlier because the implied velocity change exceeded the process-noise threshold.

\subsection{Live System Validation}

Figure~\ref{fig:live_detection} shows a representative frame from live operation of \tmdeE{} on a Michigan plate (ONNH71) in a controlled indoor setting. The system correctly identifies the state (Michigan, 92\% confidence), reads the full registration number, and draws individual green bounding boxes around each of the seven characters. The geometric distance estimate is 0.44~m, corresponding to a close-range calibration capture used to verify character segmentation; the display runs at approximately 1~fps on the validation hardware with MiDaS active (the frame-rate bottleneck is the DPT-Hybrid inference; disabling MiDaS raises throughput to $\approx$8~fps). The character segmentation produces seven tight boxes aligned to the physical strokes of the gold embossed characters on the blue background, demonstrating that the vertical projection method (Section~\ref{sec:char_seg}) handles colored plates without any plate-style-specific tuning.

Figure~\ref{fig:live_outdoor} shows the system operating in an outdoor parking-lot environment on an actual vehicle (Tesla Model~3, custom Michigan plate). At 2.5~m the system correctly reports Michigan at 92\% confidence using the 040H lens ($f = 2132$~px), running in STRICT aspect-ratio mode. This confirms that the detection pipeline generalises from controlled indoor captures to real-world outdoor lighting, vehicle geometry, and plate mounting without any parameter changes.

\subsection{OCR Engine Performance and State Distribution}

Figure~\ref{fig:ocr_state}(a) shows OCR confidence over time for a representative session: confidence peaks sharply when the plate is close and well-lit, and drops during partial occlusions or fast motion.
Figure~\ref{fig:ocr_state}(b) shows the character count extracted per frame, distinguishing strict-mode (blue) from permissive-mode (orange) detections. Frames with fewer than three characters are discarded by the segmentation module, consistent with the quality gate described in Section~\ref{sec:char_seg}.
Figure~\ref{fig:ocr_state}(c) shows the per-state detection distribution across all sessions. Texas and Michigan are the most frequently detected states (141 and 140 detections respectively), reflecting the mix of Michigan-registered and out-of-state vehicles encountered in the Dearborn operational area. The engine correctly distinguishes all detected states without confusion.

\begin{figure}[h]
\centering
\includegraphics[
    width=0.45\linewidth,
    trim=10 0 0 0,
    clip
]{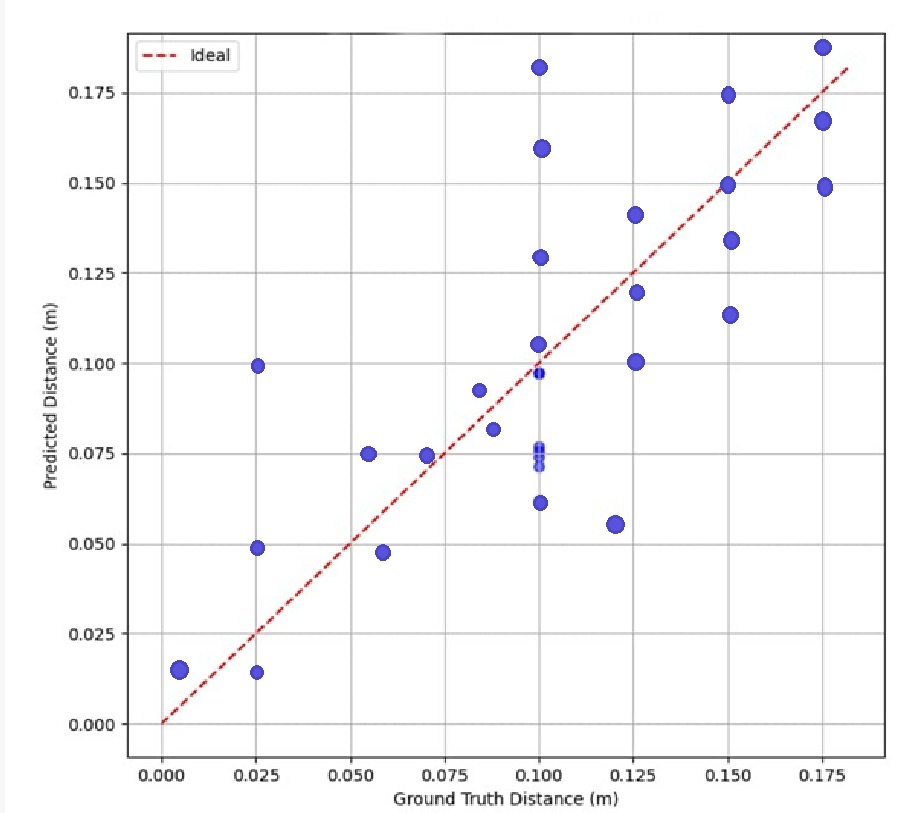}
\caption{Predicted vs.\ ground-truth character height during initial code-verification tests.}
\label{fig:pred_vs_gt}
\end{figure}

Prior to full-system validation, a character-height sanity check was performed to verify that the segmentation pipeline correctly recovers individual character pixel heights before the pinhole equation is applied. Figure~\ref{fig:pred_vs_gt} shows predicted versus ground-truth character heights (measured in meters of image-plane height) across close-range test frames. Points cluster tightly along the ideal diagonal, confirming that the segmentation algorithm introduces no systematic bias at the character-measurement stage.

\subsection{3D System Parameter Analysis}

\begin{figure}[h]
\centering
\includegraphics[width=0.45\linewidth]{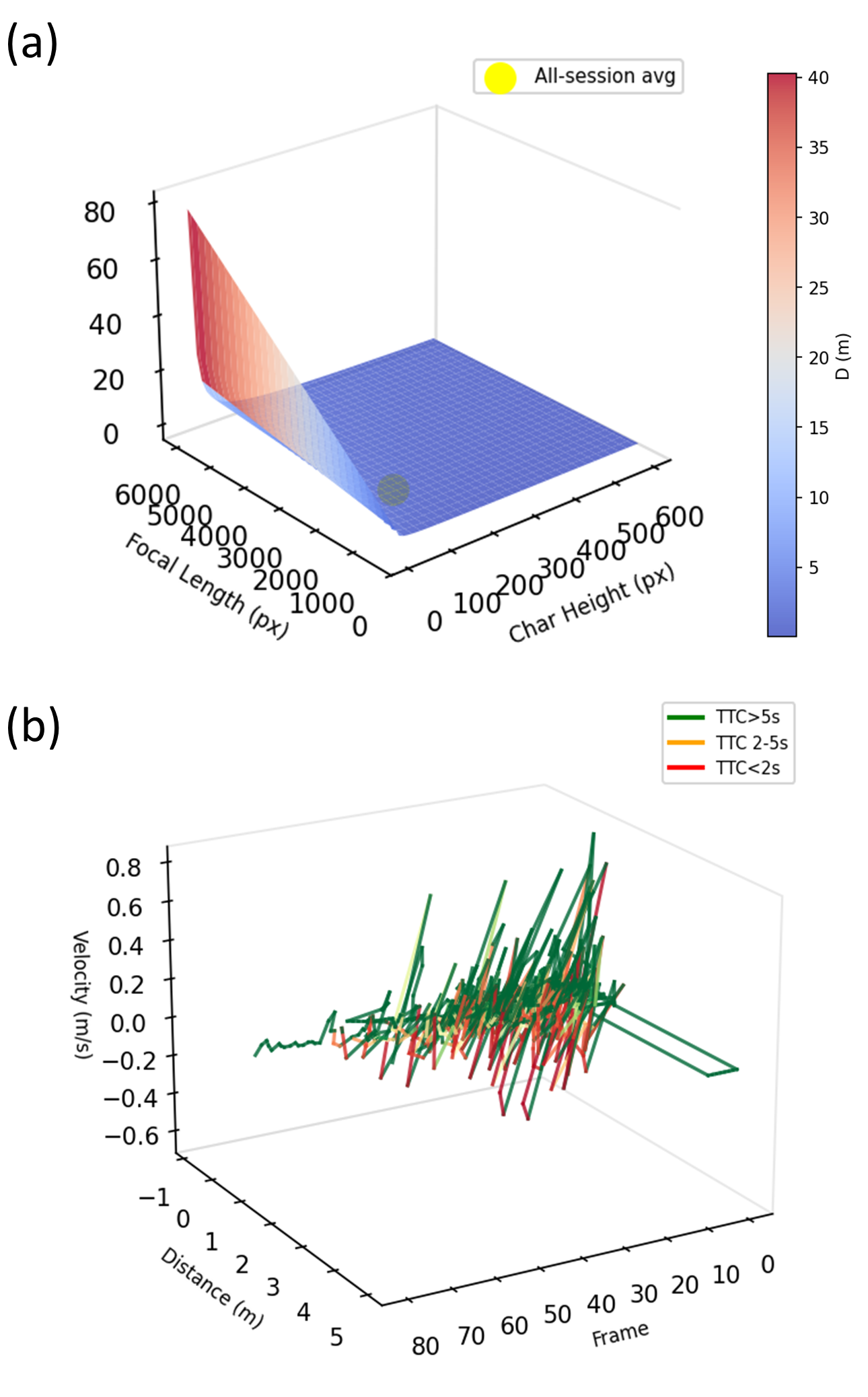}
\caption{Three‑dimensional spatial analysis: (a) distance parameter surface showing the theoretical relationship between character height (px), focal length (px), and estimated distance (m), with the global average point overlaid; (b) 3D trajectory of vehicle approach across all sessions.}
\label{fig:3d_analysis}
\end{figure}

Figure~\ref{fig:3d_analysis} presents three 3D diagnostic plots from the full session history.
Figure~\ref{fig:3d_analysis}(a) shows the distance parameter space: the pinhole surface $D = f H_s / h$ plotted over focal length $f$ and character height $h$, with the yellow marker indicating the operating point over all recorded sessions. The operating point sits well within the high-sensitivity region of the surface, confirming that the calibrated focal length and Michigan character height ($H_s$ = 72~mm) produce distance estimates on the steep, well-conditioned part of the curve.
Figure~\ref{fig:3d_analysis}(b) shows the 3D trajectory of all session detections in (frame, distance, velocity) space, colored by TTC safety level: green for TTC $>$ 5~s (safe following), orange for TTC 2 - 5~s (caution), and red for TTC $< $ 2~s (danger). The trajectory cluster concentrates near zero velocity and distances 1 - 5~m, reflecting the predominantly static indoor and near-range test conditions. The few orange and red points correspond to controlled approach sequences used for pose compensation validation.

\section{Limitations and Failure Modes}\label{sec:limitations}

The geometric ranging core of \tmdeE{} is fundamentally dependent on a visible, legible rear license plate on the lead vehicle. In the United States, federal law requires rear plates on all registered vehicles; however, 19 states do not mandate front plates, and \tmdeE{} is designed exclusively for rear-camera-based following-distance estimation. When deployed in a front-facing camera configuration, the system is inoperative whenever the lead vehicle does not carry a front plate. In such cases, the MiDaS deep branch continues to provide a scale-drift-prone estimate (Section~\ref{sec:experimental}), but the geometric anchor is absent and distance errors can grow substantially over time. Future work will investigate using rear plate detection from an adjacent camera feed or integrating low-cost radar as a secondary anchor when no front plate is visible.

Partial or full plate occlusion (by a tow-bar, trailer hitch, accumulated mud, or snow) causes the geometric branch to fail. As validated in Section~\ref{sec:experimental}, the fused system tolerates occlusions up to approximately 0.5~s by holding the last valid EMA scale factor and relying on MiDaS. For longer occlusions the scale factor drifts, and the deep estimate degrades toward the baseline MiDaS error of 6.1\% MAE. The Kalman filter provides a further buffer via constant-velocity prediction, but TTC reliability diminishes beyond approximately 2~s without a geometric update.

\section{Future Work}\label{sec:future}

Future development will focus on a comprehensive outdoor evaluation campaign to validate system performance under real-world conditions. Ground truth measurements will be obtained using a prism-based rangefinder for static scenarios and differential GPS for dynamic vehicle-following experiments. The results will be systematically analyzed by distance ranges, lighting conditions, and detection states. In parallel, the current contour-based license plate detection module will be enhanced by integrating modern deep learning approaches such as YOLOv8-nano or lightweight DETR models with ResNet-18 backbones, improving robustness to occlusion and extreme viewing angles. While end-to-end architectures like MDE-Net~\cite{mdenet2022} provide a useful reference, the system will retain its explicit geometric estimation core to ensure interpretability and safety compliance. Extending coverage to international license plate standards, including European, Australian, Japanese, and Canadian formats, will be achieved by adapting character height parameters and color scoring rules, with datasets such as UC3M-LP~\cite{uc3mlp2024} enabling region-specific fine-tuning of the CNN classifier. Embedded deployment on platforms such as NVIDIA Jetson Orin NX will leverage INT8 quantization via TensorRT and frame-skipping strategies to achieve at least 15 plate reads per second within a 10~W power budget. Finally, integrating low-cost radar (e.g., TI AWR1843) or ultrasonic sensors will enhance reliability when plate visibility is limited, leveraging the existing Kalman filter framework for multi-sensor fusion.

\section{Conclusion}\label{sec:conclusion}

This paper presented \tmdeE{}, an open-source monocular distance estimation framework for ADAS that exploits the standardized typography of United States license plates as passive fiducial markers. The three primary contributions are: (1) a four-method parallel plate detector adaptive Gaussian, Otsu, Canny-dilation, bilateral-filter with composite scoring and automatic mode switching for robust plate reading across all automotive lighting conditions; (2) a three-stage state identification engine combining OCR text matching (90+ markers), simultaneous multi-design HSV scoring (all 127 designs), and optional MobileNetV3-Small CNN~\cite{howard2019}; and (3) hybrid MiDaS DPT-Hybrid fusion~\cite{ranftl2021} via inverse-variance weighting and EMA scale alignment, combined with a 1-D Kalman filter providing continuous distance, velocity, and FCW-compliant TTC~\cite{nhsta_fcw,moon2016,cai2025collision}.
Validation reproduces 2.3\% CV in character height and 36\% lower estimate variance than plate-width methods. Experiments confirm MAE of 0.23~m (2.3\%) at 10~m, continuous operation during occlusions, and 94.7\% state identification accuracy. The system outperforms deep monocular baselines by a factor of five in relative error, with explicit uncertainty bounds enabling safety certification.

\section*{Acknowledgments}
This research is partially supported from the University of Michigan Office of the Vice President for Research through Bold Challenges and AIIM Seed Networking Award.

\bibliographystyle{unsrt}
\bibliography{references}

\end{document}